\documentclass{article}

\usepackage{arxiv}

\usepackage[usenames,dvipsnames,svgnames,table]{xcolor}

\usepackage{url} 
\usepackage{hyperref}
\usepackage{amsmath}
\usepackage{comment}
\usepackage{graphicx}
\usepackage{multirow}
\usepackage{hhline}
\usepackage{makecell}
\usepackage{graphicx}
\usepackage{booktabs} 
\usepackage{tabularx} 
\usepackage{longtable} 
\usepackage{ltablex} 
\usepackage{lscape}
\usepackage{subcaption}
\usepackage{pgfplots} 
\pgfplotsset{compat=1.18} 
\usepackage{tikz}
\usepackage{bm}

\usepackage{array}
\usepackage{multirow}
\usepackage{textcomp}
\usepackage{stfloats}
\usepackage{url}
\usepackage{verbatim}
\usepackage{graphicx}
\usepackage{ragged2e}

\usepackage{lscape}
\usepackage{longtable}
\usepackage{afterpage,array,rotating}
\usepackage{tikz}
\usetikzlibrary{positioning, fit, arrows.meta, shapes,calc}
\usetikzlibrary{shapes.geometric, arrows}
\usetikzlibrary{backgrounds} 

\tikzstyle{startstop} = [rectangle, rounded corners,
minimum width=3cm, minimum height=0.7cm,
text centered, 
text width=1.8cm, inner sep=0.5mm, font=\small,
draw=black, 
fill=red!10]

\tikzstyle{io} = [trapezium, 
trapezium stretches=true, 
trapezium left angle=70, 
trapezium right angle=110, 
minimum width=3cm, 
minimum height=1cm, text centered, 
draw=black, fill=blue!10]

\tikzstyle{process-start} = [rectangle, 
minimum width=3cm, 
minimum height=0.8cm, 
text centered, 
text width=5cm, font=\small,
draw=black, 
fill=orange!10]

\tikzstyle{process} = [rectangle, 
minimum width=3cm, 
minimum height=1cm, 
text centered, 
text width=3cm, 
draw=black, 
fill=orange!30]

\tikzstyle{decision} = [diamond, 
minimum width=3cm, 
minimum height=1cm, 
text centered, 
draw=black, 
fill=green!30]
\tikzstyle{arrow} = [thick,->,>=stealth]

\tikzstyle{comments} = [rectangle, 
minimum width=10cm, 
minimum height=0.8cm, 
text width=10cm, font=\small,
draw=none, 
fill=orange!]

\newcolumntype{P}[1]{>{\centering\arraybackslash}m{#1}}
\newcolumntype{T}[1]{>{\raggedright\arraybackslash}m{#1}}

\newcommand{\texthash}{\#}

\def\checkmark{\tikz\fill[scale=0.3](0,.35) -- (.25,0) -- (1,.7) -- (.25,.15) -- cycle;} 

\usepackage{amssymb}

\begin{document}

\title{Low-light Pedestrian Detection in Visible and Infrared Image Feeds: Issues and Challenges}

            


\author{
	Thangarajah Akilan \\
	Department of Software Engineering\\
	Lakehead University\\
	Thunder Bay, ON P7B5E1 \\
	\texttt{takilan@lakeheadu.ca} 
	\And
    Hrishikesh Vachhani\\
    Department of Computer Science\\
	Lakehead University\\
	Thunder Bay, ON P7B5E1 \\
	\texttt{hvachhan@lakeheadu.ca} 
}

\maketitle

\begin{abstract}

Pedestrian detection has become a cornerstone for several high-level tasks, including autonomous driving and video surveillance. 
There are several works focused on pedestrian detection using visible images, mainly in daytime. However, this task is very intriguing when the environmental conditions change to poor lighting or nighttime. 
Recently, new ideas have been spurred to use alternative sources, viz. Far InfraRed (FIR) temperature sensor feeds for detecting pedestrians in low-light. 
This study comprehensively reviews recent developments in low-light pedestrian detection approaches. 
It systematically categorizes and analyses various algorithms from region-based to non-region-based and graph-based learning methodologies by highlighting their methodologies, implementation issues, and challenges. 
It also outlines the key benchmark datasets that can be used for research and development of advanced pedestrian detection algorithms, particularly in low-light situations.  
\end{abstract}






\keywords{Deep learning\and low-light image processing\and machine learning\and pedestrian detection, surveillance}




\section{Introduction}\label{sec:intro}

\begin{figure*}[!bp]
\centering
\begin{tikzpicture}[
    node distance=2cm,
    every node/.style={draw, rounded corners, text width=1.5cm, align=center, font=\footnotesize, thick},
    line/.style={draw, thick, -},
    label/.style={font=\bfseries}
]

\node[rectangle, fill=orange!30, minimum width=3.4cm, text width=3.4cm, minimum height=1cm, inner xsep=0.01cm] (label1) at (-2, 2) {Hand Crafted Feature-based Models};
\node[rectangle, fill=Purple!30, minimum width=2.8cm, text width=2.8cm, minimum height=1cm, inner xsep=0.1cm] (label2) at (1.8, 2) {Region-based Models};
\node[rectangle, fill=cyan!30, minimum width=2.5cm, text width=2.5cm, minimum height=1cm] (label3) at (5.05, 2) {Non-region-based Models};
\node[rectangle, fill=red!30, minimum width=2.5cm, text width=2.5cm, minimum height=1cm] (label4) at (8.2, 2) {Graph-based Models};

\node[below=1.5cm of label1,  xshift=-1cm, yshift=0cm, inner xsep=0.01cm, fill=orange!10, font=\scriptsize] (VJ) {VJ~\cite{viola2001rapid} (2001)};
\node[above=0.3cm of VJ, xshift=0.9cm, inner xsep=0.01cm, fill=orange!10, font=\scriptsize] (HOG) {HOG~\cite{dalal2005histograms} (2005)};
\node[below=0.3cm of HOG, xshift=0.9cm, inner xsep=0.01cm, fill=orange!10, font=\scriptsize] (DPM) {DPM~\cite{felzenszwalb2008discriminatively} (2008)};

\node[below=1.5cm of label2, xshift=-0.8cm, inner xsep=0.1cm, yshift=0cm, fill=Purple!10, minimum width=1.3cm, text width=1.3cm, font=\scriptsize] (CNN) {OverFeat \cite{sermanet2013overfeat}~(2013)};
\node[above=0.35cm of CNN, xshift=0.8cm, inner xsep=0.1cm, fill=Purple!10, minimum width=1.3cm, text width=1.3cm, font=\scriptsize] (RCNN) {RCNN~\cite{girshick2014rich} (2014)};
\node[below=0.35cm of RCNN, xshift=0.8cm, inner xsep=0.1cm, fill=Purple!10, minimum width=1.3cm, text width=1.3cm, font=\scriptsize] (F-RCNN) {F-RCNN \cite{girshick2015fast}~(2015)};

\node[below=1.5cm of label3, inner xsep=0.1cm, xshift=-0.5cm, yshift=0cm, fill=cyan!10, font=\scriptsize] (YOLO) {YOLO \cite{redmon2016you} (2016)};
\node[above=0.35cm of YOLO, inner xsep=0.1cm, xshift=1cm, fill=cyan!10, font=\scriptsize] (SSD) {SSD \cite{liu2016ssd} (2016)};

\node[below=1.5cm of label4, inner xsep=0.05cm, xshift=-0.5cm, yshift=0cm, fill=red!10, font=\scriptsize]  (GNN) {Scene~GNN \cite{xu2017scene}~(2017)};
\node[above=0.35cm of GNN, xshift=1cm, inner xsep=0.05cm, yshift=0cm, fill=red!10, font=\scriptsize]  (Point-GNN) {Point-GNN \cite{shi2020point} (2020)};

\draw[line] (VJ) -- ++(0, 0.62) node[fill=orange!30, circle, scale=0.1] {} -- ++(0, 0.2);
\draw[line] (HOG.south) -- ++(0, -0.15) node[fill=orange!30, circle, scale=0.1] {} -- ++(0, -0.2);
\draw[line] (DPM) -- ++(0, 0.62) node[fill=orange!30, circle, scale=0.1] {} -- ++(0, 0.2);

\draw[line] (CNN) -- ++(0, 0.6) node[fill=Purple!10, circle, scale=0.1] {} -- ++(0, 0.2);
\draw[line] (RCNN.south) -- ++(0, -0.2) node[fill=Purple!10, circle, scale=0.1] {} -- ++(0, -0.2);
\draw[line] (F-RCNN) -- ++(0, 0.6) node[fill=Purple!10, circle, scale=0.1] {} -- ++(0, 0.2);

\draw[line] (YOLO) -- ++(0, 0.6) node[fill=cyan!10, circle, scale=0.1] {} -- ++(0, 0.2);
\draw[line] (SSD.south) -- ++(0, -0.2) node[fill=cyan!10, circle, scale=0.1] {} -- ++(0, -0.2);

\draw[line] (GNN) -- ++(0, 0.6) node[fill=red!10, circle, scale=0.1] {} -- ++(0, 0.2);

\draw[line] (Point-GNN.south) -- ++(0, -0.2) node[fill=red!10, circle, scale=0.1] {} -- ++(0, -0.2);

\draw[line width=1mm, opacity=0.2] (-3.7, 0.15) -- (9.5, 0.15);

\node[draw=Green, dashed, rounded corners, fit={(label1) (VJ) (HOG) (DPM)}, inner xsep=0.1cm, inner ysep=0.4cm] (box1) {};

\node[draw=Green, dashed, rounded corners, fit={(label2) (CNN) (RCNN) (F-RCNN) (label3) (YOLO) (SSD) (label4) (GNN) (Point-GNN)}, inner xsep=0.1cm, inner ysep=0.4cm] (box2){};

\node[font=\scriptsize, rectangle, fill=green!10, draw=white, minimum width=3.2cm, text width=3.2cm] at (box1.south) [below=-0.25cm] {Traditional Approaches};

\node[font=\scriptsize, rectangle, fill=green!10, draw=white, minimum width=4.5cm, text width=4.5cm] at (box2.south) [below=-0.25cm] {Deep Learning Approaches};

\end{tikzpicture}

\caption{Evolution of object detection methods: from traditional approaches to deep learning-inspired approaches.}\label{History}
\end{figure*}

Pedestrian detection has become a cornerstone of several vision-based applications in modern AI-assisted systems~\cite{bhatt2023dual, sundaralingam2025segattndetec}. 
It is the process of identifying human movements in input feeds from data acquisition devices, like visual cameras, and thermopile sensors, for semantic understanding of a scene. 
It is more significant than other forms of object detection since it addresses the safety concerns of the people.
Thus, it has stringent operational criteria, such as higher detection accuracy and real-time performance, which are of paramount importance to the aforesaid smart systems. 
To address this, there are several methods have been introduced as illustrated in Fig.~\ref{History} by computer vision and machine learning researchers through exploiting technological advancements that we witnessed in the past. 


\subsection{Evaluation of Pedestrian Detection}\label{sec:evluation}

The task of pedestrian detection has primarily evolved through two important eras of artificial intelligence (AI)--the traditional machine learning (ML)-based object detection era and the modern deep learning (DL)-based object detection era. 
The traditional methods arguably started when Viola and Jones (VJ)~\cite{viola2001rapid} proposed the first face detector using Haar cascade features. Later it was exploited for several other object detection tasks, including pedestrian detection. 
In this line, the Histogram of Oriented Gradient (HOG) descriptor~\cite{dalal2005histograms} was proposed by Dalal and Triggs, particularly for accurate pedestrian detection. This descriptor focuses on the objects' structural information by computing the gradient magnitude and edge direction within a certain local region of the input image. Since it creates the histogram features using both the gradient magnitude and angle, it is superior to other contemporary descriptors. 
Subsequently, Felzenszwalb~\textit{et~al.}~\cite{felzenszwalb2008discriminatively} introduced the Deformable Part Model (DPM) for object detection, which can classify pedestrians into a variety of distinct sections using a sliding window.
This algorithm has consistently produced the best detection results for several applications; however, the sliding window approach heavily burdens the computing resources as it travels through every conceivable position and window size ratio. 

In the early deep learning era, Sermanet~\textit{et al.}~\cite{sermanet2013overfeat} introduced OverFeat, a model that pioneered the use of a sliding window approach applied over feature maps learned by convolutional layers within a Convolutional Neural Network (CNN). This method enabled both classification and localization tasks in a single framework, marking one of the earliest innovations to adapt CNNs for object detection beyond traditional image classification tasks.
The success of OverFeat inspired the development of region proposal-based object detectors. For instance, the Region-based Convolutional Neural Network (R-CNN)~\cite{girshick2014rich} introduced the concept of generating region proposals as input for hierarchical feature extraction using convolutional operations, which has since influenced the structure of multi-layer neural networks for detection tasks. This region-based approach has been instrumental in advancing object detection, helping move these tasks entirely into the DL paradigm.
In RCNNs, the regions of interest (ROI) are proposed using selective search operations on which an object classier and a bounding-box regressor are trained collaboratively.  
This strategy is typically regarded as a two-stage detection framework, 
where the first stage is a feature extractor, and the second stage is the object detector. To overcome the drawbacks of RCNN, such as a rigorous training procedure and prolonged detection times, Fast-RCNN~\cite{girshick2015fast} and Faster-RCNN~\cite{ren2015faster} models were developed as improvised versions of the baseline RCNN model~\cite{9661284}. 
However, the region-based approaches still face the limitation of slow detection speed, which is not desired for real-time applications. For example, the detection speed is crucial when it comes to the task of detecting pedestrians, particularly for autonomous driving applications to prevent accidents. 
To address the limitations of region-based object detectors, the one-stage detectors, also known as non-region-based techniques, including You-Only-Look-Once (YOLO)~\cite{redmon2016you}, Single-Shot Detector (SSD)~\cite{liu2016ssd}, and RetinaNet~\cite{lin2017focal} were introduced. The object detection speed of such models is significantly increased as the object bounding box and the class probability of each region are predicted concurrently. 
In contrast to the two-stage detection frameworks, one-stage detection models instantly predict the object center and the object bounding box by placing a sequence of anchors on the feature map.

Recently, graph-based representations have gained traction in image analysis, including object detection, where they enable modeling complex relationships among objects within a scene. For example, Xu~\textit{et~al.}~\cite{xu2017scene} introduced a Scene Graph Neural Network (Scene GNN), which models objects and their relationships using graph structures. In this approach, each object instance is represented by a node characterized by its bounding box and category label, while relationships between objects are represented by directed edges linking relevant bounding boxes, thereby encoding spatial and contextual relationships within the scene. Similarly, Shi~\textit{et~al.}~\cite{shi2020point} developed Point-GNN, a GNN specifically designed for 3D object detection in LiDAR point cloud data. Point-GNN predicts object shape and class by leveraging spatial information from point cloud data, where nodes represent individual points or small groups of points, and edges capture local geometric relationships.

\subsection{Motivation}\label{sec:motivate}
In an autonomous driving environment, robust mechanisms for detecting pedestrians in a split second is a difficult task, particularly in low-light environments.
To address this, recent research works have focussed on fusion techniques for feature enhancement and the accurate detection of pedestrians. 
It motivates this study to perform a comprehensive review of fusion-based low-light image enhancement techniques, and pedestrian detection in low-light environments using visible and infrared images.
Although there are many surveys on pedestrian detection algorithms have recently been published, there are very few studies that focus on low-light or night-time pedestrian detection. In response to that, this article pragmatically investigates key pedestrian detection methodologies in low-light environments, and their fundamental implementation issues and challenges.

\subsection{Overview of the Existing Surveys}\label{sec:contribution}

In general, the majority of the existing surveys cover only certain aspects of object detection methods and do not analyze cutting-edge graph-based learning strategies, particularly for pedestrian detection in low-light conditions. 
For instance, Ahmed~\textit{et~al.} \cite{s21155116}, and Sobbahi and Tekli~\cite{ALSOBBAHI2022116848} have studied the improvement of object detection in low-light environments. 
On the other hand, Ahmed~\textit{et~al.}~\cite{ahmed2019pedestrian}, Cao~\textit{et~al.}~\cite{cao2021handcrafted}, and Xiao~\textit{et~al.}~\cite{xiao2021deep} do not study the challenges in the pedestrian detection algorithms for low-light environments. 
Among the existing surveys, Anirudh~\textit{et~al.}~\cite{bs2022low} provide an overview of low-light object detection techniques. 
That survey evaluates several low-light image enhancement approaches, including gray-scale transformation, histogram equalization, retinex-based techniques, frequency domain-based analysis, image fusion, and dehazing. 
In comparison to the traditional approaches, the authors conclude that DL-based models address the low-light picture enhancement as a residual learning problem and they are robust in improving low-light image quality. 

On the other hand, Ahmed~\textit{et~al.}~\cite{s21155116} focus on various DL-based object detection algorithms, like Faster R-CNN, Mask R-CNN, SSD, YOLO, Retina-Net, and Cascade Mask R-CNN to benchmark their computational efficiency through empirical analysis. 
Sobbahi and Tekli~\cite{ALSOBBAHI2022116848} provide a comparative study of DL-based approaches for Low-light Image (LLI) enhancement. Their study handles LLI enhancement in two strategies: (a) an offline standalone task of image enhancement, and (b) an online pre-processing phase integrated into an end-to-end training model for a high-level computer vision task. 
Nonetheless, \cite{ALSOBBAHI2022116848} does not expand the discussion to low-light pedestrian detection. 

 Dhillon and Verma~\cite{dhillon2020convolutional}
  review of various CNN architectures highlighting their characteristics focusing on three applications-- wild animal detection, small arms detection, and human detection.   
Meanwhile, the authors of another survey--\cite{ahmed2019pedestrian} review the advancements in pedestrian and cyclist recognition and intent assessment. This survey is similar to \cite{s21155116}, in terms of the algorithms, more specifically the DL models, covered in the study. However, the main difference between them is as follows. While \cite{s21155116} focuses on general object detection, \cite{ahmed2019pedestrian} aims at pedestrian and cyclist detection from the perspective of autonomous driving. 
Xaio~\textit{et~al.}~\cite{xiao2021deep} also review DL-based pedestrian detection algorithms that handle occluded and multi-scale pedestrian detection problems.  
Cao~\textit{et~al.}~\cite{cao2021handcrafted} provide an extensive review on feature selection strategies for pedestrian detection. Their study includes the models using hand-engineered features, deep features (i.e., CNN-based features), and both handcrafted and deep features.

Table~\ref{tab:existing_surveys} digests the differences between key survey studies in object/pedestrian detection and this article. 
In summary, the following are the key contributions of this article.
\begin{itemize}
    \item A taxonomical review of visible and InfraRed (IR) fusion techniques, and deep learning methods for pedestrian detection in low-light environments.
    \item A thorough summary of the low-light pedestrian detection benchmark datasets.
    \item In-depth discussion on the primary research gaps and recommendations for emerging research pathways.
\end{itemize}


\begin{figure*}[!t]
\centering
\begin{tikzpicture}[
    node distance=2.5cm,
    every node/.style={font=\scriptsize, align=center},
    rect/.style={rectangle, draw=Blue, fill=blue!10, rounded corners, minimum width=2.5cm, minimum height=1cm, inner xsep=0.01cm},
    cyl/.style={cylinder, shape border rotate=90, draw=Blue, minimum height=1cm, inner xsep=0.01cm, text width=1.8cm, minimum width=1.8cm, aspect=0.05, fill=blue!10},
    text/.style={align=left, font=\scriptsize}
]

\node (intro) [rect] {Introduction\\ Section~\ref{sec:intro}};
\node (existing) [rect, right of=intro, xshift=0.4cm] {Existing Surveys\\ Section~\ref{sec:contribution}};
\node (recent) [rect, right of=existing, xshift=0.4cm] {Model Overviews \\Section~\ref{sec:our-survey}};
\node (datasets) [cyl, right of=recent, xshift=0.1cm] {Benchmark Datasets \\ Section~\ref{sec:dataset}};
\node (conclusion) [rect, right of=datasets, xshift=0.1cm] {Conclusion \\ Section~\ref{sec:conclusion}};

\node (intro_text) [below of=intro, yshift=1.0cm, font=\tiny, align=left] {
    • Problem statement \\
    • Challenges \\
    • Contribution  \\
    • Organization
};

\node (existing_text) [ below of=existing, yshift= 1.2cm, xshift= -0.9cm, font=\tiny, align=left] {
    • Study of \\ ~~existing \\ ~~surveys
};

\node (fusion) [rect, below of=recent, yshift=1.2cm, xshift=-1.8cm, inner xsep=0.01cm, text width=1.8cm, minimum width=1.8cm] {Image fusion \\ techniques};
\node (pedestrian) [rect, below of=recent, yshift=1.2cm, xshift=0.4cm, inner xsep=0.01cm, text width=2.1cm, minimum width=2.1cm] {Pedestrian detect. techniq.};

\node (fusion_text) [ below of=fusion, yshift=1.4cm, xshift=-2.5cm, font=\tiny, align=left] {
    • Integration of RGB \& infrared images\\ ~~for feature enhancement
};
\node (pedestrian_text) [ below of=pedestrian, yshift=1.4cm, xshift=-0.5cm, font=\tiny, align=left] {
    • DL methods \\ ~~E.g., CNN, GNN
};

\node (datasets_text) [ below of=datasets, yshift=1.3cm, xshift=0.2cm, font=\tiny, align=left] {
    • Discussion of \\ ~~low-light pedestrian \\ ~~detection datasets
};

\node (conclusion_text) [below of=conclusion, yshift=0.3cm, xshift=-0.5cm, font=\tiny, align=left] {
    • Study findings \\ 
    • Highlighting best practices \\ 
    • Future direction
};

\draw[->, thick, >=Latex] (intro) -- (existing);
\draw[->, thick, >=Latex] (existing) -- (recent);
\draw[->, thick, >=Latex] (recent) -- (datasets);
\draw[->, thick, >=Latex] (datasets) -- (conclusion);

\draw[dashed] (intro) -- (intro_text);
\draw[dashed] (existing) -- (existing_text);
\draw[] (recent) -- (fusion);
\draw[] (recent) -- (pedestrian);
\draw[dashed] (fusion) -- (fusion_text);
\draw[dashed] (pedestrian) -- (pedestrian_text);
\draw[dashed] (datasets) -- (datasets_text);
\draw[dashed] (conclusion) -- (conclusion_text);

\end{tikzpicture}
\caption{The structure of this study. It comprises an introduction, a review of existing surveys, an in-depth analysis of the current research on fusion-based pedestrian detection approaches, and a detailed summary of the datasets available for low-light pedestrian detection research. Finally, it concludes the findings with future directions. }
\label{paperflow}
\end{figure*}

\setlength{\tabcolsep}{1pt}
\begin{table*}[!ht]
\centering
\caption{Comparison of existing surveys on pedestrian detection-related research.} \label{tab:existing_surveys}
\renewcommand{\arraystretch}{1.15}
\begin{tabular}{|T{3.5cm}|T{7cm}|P{0.8cm}|P{0.8cm}|P{0.8cm}|P{0.8cm}|P{0.8cm}|P{0.8cm}|P{0.8cm}|}
\hline \hline
  \centering\multirow{2}{*}{\textbf{Article}} &
  \centering\multirow{2}{*}{\textbf{Features}} & \multicolumn{7}{c|}{\textbf{The Key Topics Covered in the Article}} \\ \cline{3-9}
  & & \textbf{PD} & \textbf{DS} & \textbf{ImF} & \textbf{RnB} & \textbf{NrB} & \textbf{LLC} & \textbf{GNN}\\ \hline
Ahmed \textit{et~al.}~\cite{ahmed2019pedestrian} (2019) &
  Pedestrian and cyclist detection and intention estimation. &
  \color{black} \checkmark &
  \color{black} \checkmark &
  \color{black} \checkmark &
  \color{black} \checkmark &
  \color{black} \checkmark &
  \color{black} - &
  \color{black} - \\ \hline
Dhillon and Verma~\cite{dhillon2020convolutional} (2020) &
 Detection of wild animals, small arms, human beings, and some miscellaneous objects. &
  \color{black} - &
  \color{black} - &
  \color{black} - &
  \color{black} \checkmark &
  \color{black} - &
  \color{black} - &
  \color{black} - \\ \hline
Ahmed \textit{et~al.}~\cite{s21155116} (2021)  &
  Focuses on video, thermal, and camera-captured image-based object detection in challenging conditions. &
  \color{black} - &
  \color{black} \checkmark &
  \color{black} - &
  \color{black} \checkmark &
  \color{black} \checkmark &
  \color{black} \checkmark &
  \color{black} - \\ \hline

Cao~\textit{et~al.}~\cite{cao2021handcrafted} (2021) &
  Assesses the object detection techniques using hand-crafted features and deep features. &
  \color{black} \checkmark &
  \color{black} \checkmark &
  \color{black} - &
  \color{black} \checkmark &
  \color{black} - &
  \color{black} - &
  \color{black} - \\ \hline
Xiao~\textit{et~al.}~\cite{xiao2021deep} (2021) &
  Concentrates on occlusion and multi-scale pedestrian identification challenges &
  \color{black} \checkmark &
  \color{black} \checkmark &
  \color{black} - &
  \color{black} \checkmark &
  \color{black} \checkmark &
  \color{black} - &
  \color{black} - \\ \hline
Anirudh~\textit{et~al.}~\cite{bs2022low} (2022) &
  Overviews of several low light picture enhancing approaches &
  \color{black} - &
  \color{black} - &
  \color{black} \checkmark &
  \color{black} - &
  \color{black} - &
  \color{black} \checkmark &
  \color{black} - \\ \hline
Sobbahi and Tekli~\cite{ALSOBBAHI2022116848} (2022) &
  Low light image enhancement and object detection &
  \color{black} - &
  \color{black} - &
  \color{black} \checkmark &
  \color{black} \checkmark &
  \color{black} \checkmark &
  \color{black} \checkmark &
  \color{black} - \\ \hline

This survey & Thermal and visible image fusion and deep learning techniques for low-light pedestrian detection &
  \color{black} \checkmark &
  \color{black} \checkmark &
  \color{black} \checkmark &
  \color{black} \checkmark &
  \color{black} \checkmark &
  \color{black} \checkmark &
  \color{black} \checkmark \\ \hline\hline
\end{tabular}
{\footnotesize Note: PD - Pedestrian detection, DS - Datasets, ImF - Image fusion, RnB - Region-based, NrB - Non-region-based, LLC - Low light condition, GNN - Graph neural network. ``\checkmark'' and ``-" denote that the topic is covered and not covered in the corresponding survey,  respectively.}
\end{table*}

\subsection{Organization}\label{sec:org}

\begin{figure}[!tp]
    \centering
    \includegraphics[width=\linewidth]{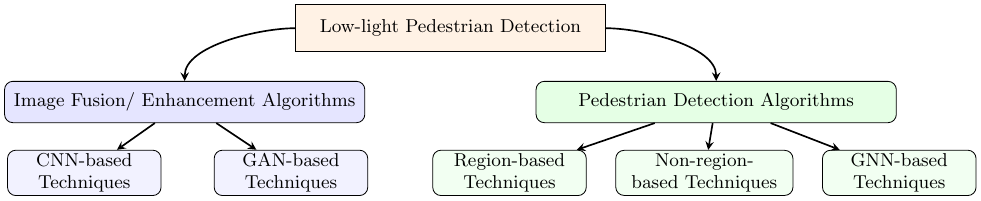}

 \caption{The taxonomy of image fusion and low-light pedestrian detection methods.}
\label{taxonamy}
\end{figure}

As summarized in Fig.~\ref{paperflow}, Section~\ref{sec:intro} lays the foundation and the concepts of pedestrian detection, and demonstrates the significance of this study.
Section~\ref{sec:our-survey} meticulously analyzes the advanced pedestrian detection algorithms and their challenges. It is divided into two subsections---input enhancement using fusion strategies and pedestrian detection models. The subsection on input enhancement examines several RGB and IR data fusion methods and their challenges. 
It is further categorized into CNN-based, and Graph Attention Network (GAN)-based fusion schemes.
The subsection on pedestrian detection models focuses on deep-learning methodologies for detecting pedestrians in low-light conditions. This subsection is divided into several constituents, viz. region-based, non-region-based, and Graph Neural Network (GNN). Fig.~\ref{taxonamy} illustrates the categorization of both image fusion and pedestrian detection techniques, together with their sub-divisions through a tree diagram. Section~\ref{sec:dataset} focuses on the benchmark datasets widely used in pedestrian detection-related research.
A discussion of the research gaps along with concluding remarks is presented in Section~\ref{sec:conclusion}.

\newpage

\section{Pedestrian Detection in Low-Light Environment}\label{sec:our-survey}

\subsection{Information Fusion Techniques for Image Enhancement}


\begin{figure}[!t]
    \centering
\begin{tikzpicture}
    \node (irImg){\includegraphics[width=4.5cm]{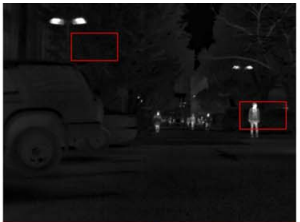}};
    \node [below=-0.15cm of irImg] (irText) {IR Image};
    \node [below=-3.4cm of irImg, xshift=-0.8cm, text=red] (roiA) {$A_{ir}$};
    \node [below=-2.4cm of irImg, xshift=1.7cm, text=red] (roiB) {$B_{ir}$};
    \node [below=0.2cm of irImg] (rgbImg) {\includegraphics[width=4.5cm]{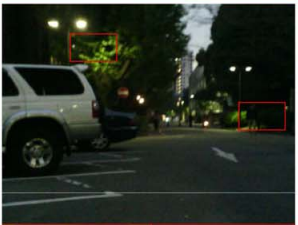}};
    \node [below=-0.1cm of rgbImg] (irText) {Visible Image};
    \node [below=-3.45cm of rgbImg, xshift=-0.8cm, text=red] (roiA) {$A_{vi}$};
    \node [below=-2.45cm of rgbImg, xshift=1.7cm, text=red] (roiB) {$B_{vi}$};
    \node (fe) [startstop, right=of irImg, yshift=0.4cm,  xshift=-0.5cm, text width=1.8cm, minimum width=1.8cm, minimum height=1.5cm, fill=blue!10] {Feature Extraction};
    \node (fuse) [startstop, right=of fe, yshift=0cm,  xshift=-0.5cm, text width=1.8cm, minimum width=1.8cm, minimum height=1.5cm, fill=blue!10] {Fusion Strategy};
    \node (rule) [below=of fuse, yshift=0.5cm,  xshift=0cm, text width=4cm, minimum width=4cm, minimum height=1.5cm, fill=white!10, font=\footnotesize] {Fusion rules, ML models, \newline or DL models};
    \node (recon) [startstop, right=of fuse, yshift=0cm,  xshift=-0.5cm, text width=3cm, minimum width=3cm, minimum height=1.5cm, text centered, fill=blue!10] {Input \\Reconstruction};
    \node [below= of recon, xshift=-1.4cm, yshift=-1.4cm] (output) {\includegraphics[trim={0.1cm, 0cm, 0cm, 0cm}, clip, width=4.5cm]{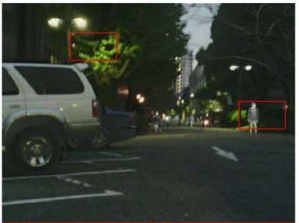}};
    \node [below=-3.5cm of output, xshift=-0.8cm, text=red] (roiA) {$A_{re}$};
    \node [below=-2.45cm of output, xshift=1.7cm, text=red] (roiB) {$B_{re}$};
    \begin{scope}[on background layer]
        \node[draw=blue, draw opacity=0.99, fill=blue, fill opacity=0.05, thick, rounded corners, fit=(irImg) (rgbImg) (irText), inner sep=0mm](bb){};        
    \end{scope}

    \draw [->] (bb.east) -| ++(0.2cm, 0) |- (fe.west);
    \draw [->] (fe.east) to (fuse.west);
    \draw [dashed] (fuse.south) to (rule.north);
    \draw [->] (fuse.east) to  (recon.west);
    \draw [->] (recon.east) -| ++(0.4cm, 0) |- (output.east);

\end{tikzpicture}
\caption{A conceptual diagram of information fusion methods used in low-light pedestrian detection using IR and regular visible (RGB) images. $A_{ir}$ and $B_{ir}$ are two RoIs in the IR image and their corresponding RoIs in the RGB image are annotated as $A_{vi}$ and $B_{vi}$, respectively. The corresponding RoIs -- $A_{re}$ and $B_{re}$ in the output image have improved these regions' information. The sample images in this figure are adopted from \cite{10006826}.}
\label{Conseptional diagram}
\end{figure}

Fig.~\ref{Conseptional diagram} depicts the concepts of fusion techniques generally employed for feature enhancement in low-light pedestrian detection. It subsumes three stages--(i) a feature extractor that extracts features from the source inputs, (ii) an encoder that takes features from the two streams of visible and infrared (IR) data, and (iii) a reconstructor that combines encoded information into a final feature enhanced image. 
A general overview of the performance and limitations of existing fusion methodologies are compared and summarized in Table~\ref{tab:FusionTable} on page \pageref{tab:FusionTable}. 



\onecolumn
{\small
\begin{landscape}
\setlength{\tabcolsep}{3pt}
\begin{longtable}[c]{T{2cm}T{1.5cm}T{4.0cm}T{4.5cm}T{3.7cm}T{5.5cm}}
\caption{The approaches used for fusion of visible and IR images. Note: Accu. - Accuracy, AP - Avg. precision, AG - Avg. gradient, CC - Correlation coefficient, EN - Engtropy, mAP - Mean Avg. precision, FPS - Frame per second, Log AMR - Log Avg. miss rate, mIoU - Mean intersection over union, SF - Standard deviation and fidelity, SNR - Signal to noise ratio.}
\label{tab:FusionTable}\\
\hline 
\textbf{Authors} &
  \centering\textbf{Dataset} &
  \centering\textbf{Methodology} &
  \centering\textbf{Merits} &
  \centering\textbf{Best Performance} &
  \hspace{1cm}\textbf{Limitations}  \\
  \hline
 \endfirsthead

 \hline
 \multicolumn{6}{c}{Continuation of Table \ref{tab:FusionTable}}\\
 \hline
 \textbf{Authors} &
  \centering\textbf{Dataset} &
  \centering\textbf{Methodology} &
  \centering\textbf{Merits} &
  \centering\textbf{Best Performance} &
  \hspace{1cm}\textbf{Limitations}\\
  \hline
 \endhead

 \hline
 \endfoot

 \hline
 \multicolumn{6}{c}{End of Table \ref{tab:FusionTable}}\\
 \hline\hline
 \endlastfoot
Golcare-narenji \textit{et~al.} \cite{golcarenarenji2022illumination} &   KAIST & Enhanced correlation coefficient (ECC) and arithmetic image addition &
  Reduction in computing makes it feasible for portable, and resource-constrained platforms & 95.01\% mAP, 42.2 FPS inference seeped &  Focused on the reduction of computational complexity; but no testing is conducted on actual embedded devices \\ \hline
Song \textit{et~al.} \cite{song2021multispectral} &  KAIST & Multi-spectral feature fusion network (MSFFN) &  Handling tiny targets of various sizes without increasing the computations &  85.4\% accu., 56 FPS inference speed & Failing to detect pedestrians in the presence of occlusion \\ \hline
Deng \textit{et~al.} \cite{deng2021pedestrian} &  KAIST &  Multi-layer fusion network with faster R-CNN & Using combination of RGB and infrared data  & 91.2\% AP, 0.14 FPS & Ignored the impact of illumination changes \\\hline
Kim \textit{et~al.} \cite{kim2018pedestrian} &  KAIST, Caltech &  Faster R-CNN & Object detection performance is enhanced using frame sequence information &  Outperforms several SOTA methods in detection by over 5\%. & Inference speed is not considered. Poor performance in foggy, rainy, or snowy weather conditions. Heavy reliance on image preprocessing. \\ \hline
Li \textit{et~al.} \cite{li2018infrared} & TNO &  Infrared and visible image fusion using latent low-rank representation & Optimized model with acceptable performance & Reconstructed image after fusion looks more natural with 0.01596 SNR &  The model is not tested for a high-level task, like pedestrian detection \\\hline
Rashed \textit{et~al.} \cite{rashed2019fusemodnet} &  KITTI-Dark & CNN-based moving object detection with fusion of RGB and LiDAR data & Performed better regardless of the lighting conditions &  71.2\% mIoU, 18 FPS &  The cost of data fusion (RGB + rgbFlow + lidarFlow) is higher due to the increased number of weights in the encoder \\\hline
Pei \textit{et~al.} \cite{pei2020fast} & KAIST multi-spectral pedestrian &  RetinaNet, Feature pyramid network (FPN) &  Aggregating multi-scale feature output maps of the FPN for detecting different size targets & 27.6\% Log AMR, 0.129 PFS &  It requires more experimental analysis and performance evaluations \\ \hline
Jian \textit{et~al.} \cite{jian2020sedrfuse} &  KAIST, FLIR, and TNO &  Symmetric encoder-decoder network with residual block fusion &  Excellent fusion quality & 4.43 AG, 6.98 EN, 16.25 SF & Computational complexity and resource requirements.
  Sometimes the fusion loses some useful details and produces unacceptable artifacts\\ \hline
  Gao \textit{et~al.} \cite{gao2022dcdr} & FLIR and TNO &
  Densely connected disentangled representation GAN (DCDR-GAN) & A novel optimization objective incorporated both the content and modal reconstruction losses. &
  The maximum rise in objective index scores is 9.28\%. &
  The moderate time complexity of 8.64 seconds. \\\hline
Tang~\textit{et~al.} \cite{tang2022piafusion} &
  Multi-Spectral Road Scenarios (MSRS) &
  End-to-end CNN-based framework as a backbone. &
  CMDAF + Halfway fusion - Merged the complimentary and common information in an adaptable manner depending on the illumination conditions. &
  Generated pleasant fused outputs and transmitted more data from the source photos to fused images. Achieved the top spot on the Qabf metric &
  Relatively time-consuming. \\ \hline
Wagner~\textit{et~al.} \cite{wagner2016multispectral} &
  KAIST Multi-spectral Pedestrian Detection &
  Deep Fusion Architectures - Early and Late Fusion &
  Compared each fusion architecture's performance both with and without pre-training. &  Performance of LateFusion with pre-training architecture improved by 5.12\% during the day and by 10.4\% at night. compared to the baseline. &
  - \\ \hline
Zhang~\textit{et~al.} \cite{zhang2021gan} &   TNO and RoadScene &
  GAN-FM: GAN with Full-Scale Skip Connection and Dual Markovian Discriminators &  Prevention of background texture because of the edge diffusion of the high-contrast target achieved by efficient joint gradient loss. &   Achieved the largest average values on SF, MI, VIF and SD, out of six quantitative assessment metrics. &   Run time was comparatively higher. \\\hline
Yang~\textit{et~al.} \cite{yang2021infrared} &  TNO and RoadScene &
  Texture Conditional Generative Adversarial Network (TC-GAN) & Texture map + Adaptive guided filter were used in the proposed multiple decision map-based fusion technique to extract significant texture data from the input images. &
Produced superior fusion outcomes than the other advanced fusion methods, in terms of subjective and objective assessments. & In the context of the IVIF task, only a portion of the data from the infrared and RGB image models were typically complimentary. \\ \hline
Li~\textit{et~al.} \cite{li2020multigrained} &
  TNO and BEMP &
  Multi-grained Attention module into ED network to Fuse (MgAN-Fuse) &
  An extra feature loss function was built to capture significant elements of the RGB image. &
  Outperformed VIFF, SD, and EN metrics while maintaining comparable performance on MI. &
  Not computational cost efficient as other systems. \\ \hline
  Liao~\textit{et~al.} \cite{liao2022cross} &
  VT821, VT1000, VT5000 &
  Cross-Collaborative Fusion-Encoder network (CCFENet) &
  Integrated multi-modal data to efficiently identify tiny objects in the complex highlighted area and muted the background noise. &
  Performed best in terms of FPS, Size, and FLOPs. &
  - \\
& & & & & \\ \hline
Li~\textit{et~al.} \cite{li2022mafusion} &
  MS-COCO and TNO &  Multiscale Attention network for infrared and RGB image fusion (MAFusion) &
  To reduce the loss of critical data and avoid inconsistencies, the details and semantics information at each level were fully fed into the fused images. &
  Performed best in QNCIE, FMI, QW, QP, SD, QC B, and QC evaluation metrics. & -
   \\ \hline
Wang~\textit{et~al.} \cite{wang2022unidirectional} &
  VT5000, VT821, VT1000 &
  LDFM (Local Detail Driven Fusion Module), Encoder-Decoder Module &
  Effectively worked on proper object detection effects and sharpened boundaries to challenging scenarios such as multiple salient and small salient objects, or background clutter. &
  On VT821, weighted F-measure and Adaptive F-measure improved by 4.3\% and 2.7\% respectively. &
  Had trouble correctly categorizing the surrounding region as background. \\ \hline
Chen~\textit{et~al.} \cite{chen2020pedestrian} &
  KMU and CVC-09 &
  CNN based encoder-decoder &
  Fast computation speed. Eliminated the background interference well which enabled better low light/ night-time pedestrian detection. &
  Obtained the highest AP of 97.5\%. The computation speed is 0.03s per image. &
  Performance degradation due to occlusions. Obtained false detections among the crowd. \\ \hline
Wang~\textit{et~al.} \cite{wang2022res2fusion} &
  TNO and Roadscene &
  Res2Fusion - Dense Res2net and double non-local attention models &
  Attention models enhanced feature maps acquired by the encoder network to focus on salient infrared objects and distinguishable visual features. &
  Nine evaluation metrics: EN, NCIE, SD, MI, FMI, SCD, Qabf, MS-SSIM, and VIF. Earned the best metrics for NCIE, MI, FMI, Qabf, and VIF. &
  Computational complexity was significant, limiting its practical applicability. \\ \hline
 
\end{longtable}
\end{landscape}
}


\subsubsection{CNN-based Fusion}

\begin{figure}[!t]
    \centering
    \includegraphics[width=\linewidth]{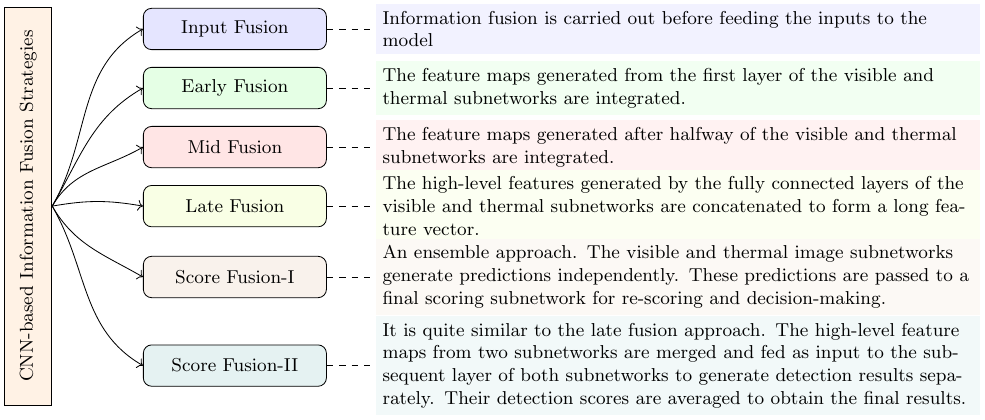}
\caption{An overview of CNN-based information fusion strategies used in object detection. Refer to Fig.~2 in ~\cite{li2019illumination} for an architectural description of these fusion approaches.}
\label{Fusiontechniques}
\end{figure}

There are six CNN-based fusion techniques, namely input fusion, early fusion, halfway fusion, late fusion, score fusion-I, and score fusion-II, each combining color and thermal modalities at stages of the network (finer visual information at the bottom layers and greater semantic interpretations at the top layers), as demonstrated in Fig.~\ref{Fusiontechniques}.

\begin{figure}[!ht]
\centering
\begin{tikzpicture}[
    block/.style = {rectangle, draw, rounded corners, text width=4cm, align=center, minimum height=1.5cm},
    arrow/.style = {thick,->,>=stealth}
]

\node (visible_image) [block, minimum width=3.0cm, minimum height=1cm, text width=3.0cm, inner sep=1pt, font=\footnotesize] {RGB image};
\node (thermal_image) [block, right=4.5cm of visible_image,  minimum width=3.0cm, minimum height=1cm, text width=3.0cm, inner sep=1pt, font=\footnotesize] {IR image};

\node (resnet_visible) [block, below=0.8cm of visible_image, yshift=0.5cm, minimum width=3.0cm, minimum height=0.8cm, text width=3.0cm, inner sep=1pt] {{\footnotesize ResNet50}\\ {\scriptsize (Feature Learning from RGB Image)}};
\node (resnet_thermal) [block, below=0.8cm of thermal_image, yshift=0.5cm, minimum width=3.0cm, minimum height=1cm, text width=3.0cm, inner sep=1pt] {{\footnotesize ResNet50}\\ {\scriptsize (Feature Learning from IR Image)}};

\node (resnet_visible_fpn) [block, below=0.8cm of resnet_visible, yshift=0.5cm, minimum width=3.0cm, minimum height=1cm, text width=3.0cm, inner sep=1pt, ] {{\footnotesize FPN}};
\node (resnet_thermal_fpn) [block, below=0.8cm of resnet_thermal, yshift=0.5cm, minimum width=3.0cm, minimum height=1cm, text width=3.0cm, inner sep=1pt] {{\footnotesize FPN}};

\node (concat) [block, right=of resnet_visible_fpn, xshift=-.3cm, minimum width=3.0cm, minimum height=1cm, text width=3.0cm, inner sep=1pt, font=\footnotesize] {Feature \\ Concatenation\\(4 stages)};

\node (rpn) [block, below=of concat,, xshift=0cm, minimum height=1cm, text width=3cm, minimum width=3.0cm, inner sep=.1pt, font=\footnotesize] {RPN};
\node (object_detection) [block, right=of rpn, xshift=-0.2cm, minimum width=3.0cm, minimum height=1cm, text width=3.0cm, inner sep=1pt, font=\footnotesize] {Object Detection \& Classification};

\draw[arrow] (visible_image) -- (resnet_visible);
\draw[arrow] (thermal_image) -- (resnet_thermal);

\draw[arrow] (resnet_visible) -- (resnet_visible_fpn);
\draw[arrow] (resnet_thermal) -- (resnet_thermal_fpn);

\draw[arrow, ultra thick] (resnet_visible_fpn) -- (concat);
\draw[arrow, ultra thick] (resnet_thermal_fpn) -- (concat);

\draw[arrow] (concat) -- node[anchor=east, xshift=0.1cm, minimum width=3.8cm, minimum height=1cm, text width=3.8cm, inner sep=1pt, font=\scriptsize, gray] {Region proposals are made at each level of FPN feature map after concatenation} (rpn);

\draw[arrow] (rpn) -- (object_detection);
\end{tikzpicture}

\caption{An overview of a CNN-based fusion introduced in \cite{pei2020fast}. The subnetworks used for RGB and IR streams are identical to each other.}
\label{fig:RetinaNet-fusion}
\end{figure}

As summarized in Table~\ref{tab:FusionTable}, Pei~\textit{et~al.}~\cite{pei2020fast} develop a novel architecture for the sole purpose of detecting as many pedestrians as possible with a fusion of RGB and IR images as shown in Fig.~\ref{fig:RetinaNet-fusion}. Therefore, the authors perform empirical studies on their proposed structure with two types of fusion stages, such as early fusion (occurs before any feature learning), and late fusion (occurs after the Feature Pyramid Network (FPN), and various fusion strategies, like liner concatenation (stacking up the two feature maps at the same spatial locations), applying mathematical operators, including sum, max, and
average on the two feature maps at the same spatial location.
Their study reveals that the late fusion with feature approach can achieve the highest detection rate. The reason behind this is as follows: (i) the early-fusion model is incapable of capturing key multi-modal information, (ii) the late-fusion technique by an arithmetic summation of the two weak feature maps of the FPN outputs creates a stronger multi-spectral feature map that embeds multi-scale object information, resulting in accurate detection. 
The comparative analysis shows that the proposed model achieves the lowest miss rate and excellent pedestrian detection in half the run time of the state-of-the-art frameworks. 

\begin{figure*}[!tp]
\centering
\begin{tikzpicture}[
    block/.style = {rectangle, draw, rounded corners, text width=4cm, align=center, minimum height=1.5cm},
    arrow/.style = {thick,->,>=stealth}
]

\node (visible_image) [block, minimum width=3.0cm, minimum height=1cm, text width=3.0cm, inner sep=1pt, font=\footnotesize] {Visible image};
\node (thermal_image) [block, right=4.5cm of visible_image,  minimum width=3.0cm, minimum height=1cm, text width=3.0cm, inner sep=1pt, font=\footnotesize] {Thermal image};

\node (resnet_visible) [block, below=0.8cm of visible_image, yshift=0.5cm, minimum width=3.0cm, minimum height=1cm, text width=3.0cm, inner sep=1pt, ] {{\footnotesize ResNet50}\\ {\scriptsize (Feature Learning from Visible Image)}};
\node (resnet_thermal) [block, below=0.8cm of thermal_image, yshift=0.5cm, minimum width=3.0cm, minimum height=1cm, text width=3.0cm, inner sep=1pt] {{\footnotesize ResNet50}\\ {\scriptsize (Feature Learning from Thermal Image)}};

\node (concat) [block, right=of resnet_visible, xshift=-.3cm, minimum width=3.0cm, minimum height=1cm, text width=3.0cm, inner sep=1pt, font=\footnotesize] {Feature \\ Concatenation\\(4 stages)};
\node (fpn) [block, below=0.8cm of concat, minimum width=1cm, minimum height=0.8cm, text width=1cm, inner sep=.1pt, font=\footnotesize] {FPN};
\node (rpn) [block, right=of fpn, xshift=-.65cm, minimum width=1.0cm, minimum height=0.8cm, text width=1cm, inner sep=.1pt, font=\footnotesize] {RPN};
\node (object_detection) [block, right=of rpn, xshift=-.65cm, minimum width=3.0cm, minimum height=1cm, text width=3.0cm, inner sep=1pt, font=\footnotesize] {Object Detection \& Classification};

\draw[arrow] (visible_image) -- (resnet_visible);
\draw[arrow] (thermal_image) -- (resnet_thermal);

\draw[arrow, ultra thick] (resnet_visible) -- (concat);
\draw[arrow, ultra thick] (resnet_thermal) -- (concat);

\draw[arrow] (concat) -- node[anchor=east, xshift=0.1cm, minimum width=2.5cm, minimum height=1cm, text width=2.5cm, inner sep=1pt, font=\scriptsize, gray] {Features from the top concatenation} (fpn);
\draw[arrow] (fpn) -- (rpn);
\draw[arrow] (rpn) -- (object_detection);

\end{tikzpicture}

\caption{A high-level overview of a CNN-based multilayer fusion model proposed in \cite{deng2021pedestrian}. Note: FPN - feature pyramid network, RPN - region proposal network.}
\label{fig:MLF-FRCN}
\end{figure*}

\begin{figure*}[!tp]
\centering
\begin{tikzpicture}[
    font=\scriptsize,
    block/.style={rectangle, draw, minimum width=1.8cm, minimum height=1.2cm, align=center},
    encoder/.style={trapezium, draw, shape border rotate=270, minimum width=1.6cm, minimum height=1.0cm, inner sep=0.05cm, align=center},
    fusion/.style={rectangle, draw, minimum width=2.5cm, minimum height=2.5cm, align=center},
    decoder/.style={trapezium, draw, inner sep=0.05cm, shape border rotate=90, minimum width=1.6cm, minimum height=1.0cm, align=center},
    arrow/.style={->, thick, >=stealth},
]

\node[block, minimum height=2cm, minimum width=2cm, text width=2cm] (ir) {IR\\Image};
\node[block, below=0.2cm of ir, minimum height=2cm, minimum width=2cm, text width=2cm] (rgb) {RGB\\Image};
\node[encoder, right=1cm of ir] (encoder_ir) {Encoder\\Subnet};
\node[encoder, right=1cm of rgb] (encoder_rgb) {Encoder\\Subnet};
\node[fusion, right=1.5cm of $(encoder_ir)!0.5!(encoder_rgb)$] (fusion) {2-stage\\Feature Fusion\\with Attention};
\node[decoder, right=0.8cm of fusion] (decoder) {Decoder\\Subnet};
\node[block, right=0.8cm of decoder, minimum height=2cm, minimum width=2cm, text width=2cm] (fused) {Reconstructed\\Fused Image};
\draw[arrow] (ir) -- (encoder_ir);
\draw[arrow] (rgb) -- (encoder_rgb);
\draw[arrow] (encoder_ir) -- (fusion);
\draw[arrow] (encoder_rgb) -- (fusion);
\draw[arrow] (fusion) -- (decoder);
\draw[arrow] (decoder) -- (fused);
\draw[arrow, thick] ($(encoder_ir.north) + (0,0)$) -- ++(0,0.3) -| ($(decoder.north)+(0,0.3)$) -- (decoder.north);
\draw[arrow, thick] ($(encoder_rgb.south) + (0,0)$) -- ++(0,-0.3) -| ($(decoder.south)+(0,-0.3)$) -- (decoder.south);
\node[above=0.4cm of $(encoder_ir.north)!0.5!(decoder.north)$, text=gray] {3-level feature recovery paths};
\node[below=0.4cm of $(encoder_rgb.south)!0.5!(decoder.south)$, text=gray] {3-level feature recovery paths};
\end{tikzpicture}
\caption{An illustration of the UNet-like fusion approach presented in \cite{jian2020sedrfuse}.}
\label{fig:SEDRFFuse}
\end{figure*}

Similar to \cite{pei2020fast}, Deng~\textit{et~al}~\cite{deng2021pedestrian} also come up with a multi-layer fusion (MLF) framework as shown in Fig.~\ref{fig:MLF-FRCN} for low-light pedestrian recognition named MLF-FRCNN. In this method, feature maps from the RGB and IR channels are concatenated before applying the FPN in contract to \cite{pei2020fast}, where the multi-view feature maps from RGB and IR feeds are concatenated in the FPN. 
The authors conduct an ablation study to compare the performances of MLF-FRCNN using uni-modality RGB input, uni-modality IR input, halfway fusion, and full MLF under various lighting circumstances.

Song~\textit{et~al.}~\cite{song2021multispectral} introduce a reliable Multi-Spectral Feature Fusion Network (MSFFN) that fully combines data taken from IR and regular RGB light channels. Their model uses an improved version of YOLOv3 network to fuse multi-scale semantic features extracted by two modules--multi-scale feature extractors of visible images and IR images for robust pedestrian recognition. 
The experimental study on a benchmark dataset, KAIST (cf.~Section~\ref{sec:dataset}) shows that the MSFFN model performs better than single input modality-based pedestrian detection in different scenarios, including occlusion, various object sizes, and time variations (day or night). 
Similarly, Rashed~\textit{et~al.}~\cite{rashed2019fusemodnet} develop a fusion architecture called Fusemodnet that embeds camera sensory and LiDAR information to address the moving object detection problem under low-light conditions for autonomous driving.

Li and Wu~\cite{li2018infrared}, on the other hand, perform the fusion of visible and IR images using a statistical approach--Latent Low-Rank Representation (LatLRR). In which, the source images are decomposed into low-rank and salient features, which contain global and local structural information, respectively. The fusion process uses a weighted average operator to combine the low-rank components of the sources. For the salient component, it uses a sum operator for integrating the sources. 
Likewise, the authors of \cite{golcarenarenji2022illumination}, propose an image registration and fusion model called BlendNet AI to deal with the difficulties of detecting people from Unmanned Aerial Vehicles (UAVs) at great heights and different viewing angles. 
 Hence, the proposed methodology in \cite{golcarenarenji2022illumination} uses Enhanced Correlation Coefficient (ECC) and arithmetic image addition with customized weights. It overcomes the challenges, like poor illumination and complex background using optical and thermal imagery captured by the UAVs.  
Inspired by the faster R-CNN and RPN architectures König~\textit{et~al.}~\cite{konig2017fully} investigate a multi-spectral image fusion CNN using three visual optical (VIS) spectrum channels and a thermal IR channel. They tailor the VGG-16~\cite{simonyan2014very} to generate multispectral features for the RPN and perform halfway-fusion as in \cite{liu2016multispectral}.   
They analyze the person detection performance of the fusion using feature map concatenation and Network-In-Network (NIN) at the first five Convolutional (Conv) layers of the baseline network. Their experimental study shows the proposed method with fusion at the third Conv layer and ten proposals per input achieve an acceptable miss rate as compared to other existing methods at the time.

\begin{figure*}[!tp]
\centering
\includegraphics[trim={0.2cm, 0cm, 0.2cm, 0.1cm}, clip, width=0.99\linewidth]{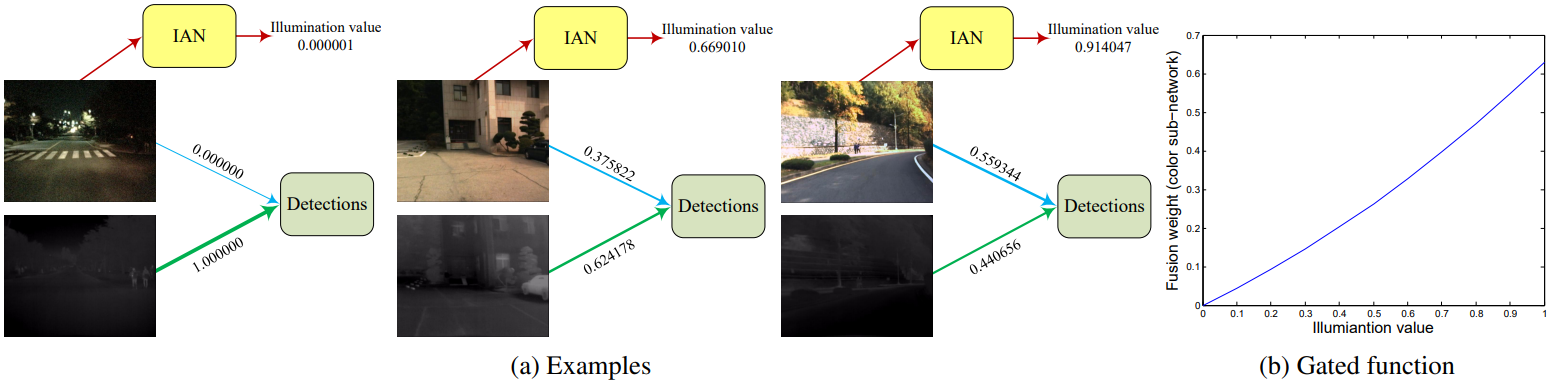}
\caption{Illustration of illumination-aware detection model and its gated function, i.e., weighting scheme (adapted from \cite{li2019illumination}). The illumination-aware weighting scheme adaptively estimates the weights for RGB and IR image learning subnetworks. Note: IAN - Illumination aware shallow network that estimates the illumination condition.}
\label{fig:IANFuse}
\end{figure*}

Jinan~\textit{et~al.}~\cite{jian2020sedrfuse}, on the other hand, use a UNet-like fusion approach that fuses encoded bottleneck feature maps of the independent input modalities--IR and RGB through an attention-based mechanism, and progressively fuses the decoded feature maps using residual connections as shown in Fig.~\ref{fig:SEDRFFuse} on page \pageref{fig:SEDRFFuse}.
This symmetric encoder-decoder fusion with the residual block is named SEDRFuse. Although the performance of this method is comparable to similar methodologies, this model is reported to have higher computational efficiency. 

Contrary to the above works, Li~\textit{et~al.}~\cite{li2019illumination} focus on building an illumination-aware fusion model using VGG-16 and RPN. Firstly, they examine the six fusion strategies elaborated earlier in Fig.~\ref{Fusiontechniques} and propose a weighting scheme based on illumination conditions to integrate color and thermal modalities. An example of the weighted fusion is shown in Fig.~\ref{fig:IANFuse}. 
Their ablation study shows that illumination conditions influence the degree of confidence in pedestrian detection. However, the experimental findings do not show any conclusive evidence about the proposed model's effectiveness in nighttime scenarios since the thermal modality alone achieves the lowest log-average miss rate.

\subsubsection{GAN-based Fusion}
\begin{figure}[!ht]
\centering
\includegraphics[trim={0.2cm, 0.2cm, 0.1cm, 0.1cm}, clip, width=0.6\linewidth]{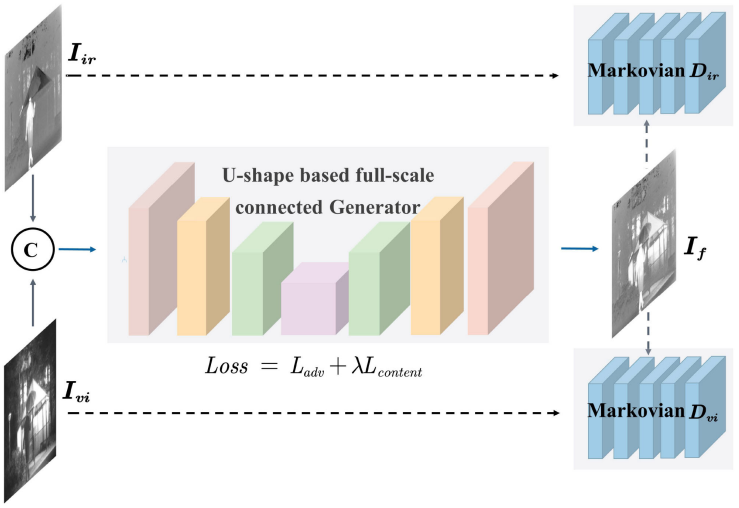} 
\caption{Illustration of the GAN w/ dual Markovian discriminators, aka. GAN-FM modeled in \cite{zhang2021gan}. $\mathcal{I}_{ir}$ - IR image, $\mathcal{I}_{vi}$ - visible image, $\mathcal{I}_{f}$ - fused image, $\mathcal{L}_{adv}$ - adversarial loss, and $\lambda\mathcal{L}_{content}$ - weighted content (the meaningful information in source images) loss.  \footnotesize Note: The block diagram is adopted as it is from \cite{zhang2021gan}.}
\label{fig:GAN-FM}
\end{figure}

Generative Adversarial Networks (GANs) have been increasingly deployed for infrared and regular image fusion. For example, to completely retain effective information in infrared and visible images, the authors of \cite{zhang2021gan} exploit GAN and Markovian discriminators (cf.~Fig.~\ref{fig:GAN-FM}). While it maintains the high-contrast target, it retains the crucial background texture and eliminates the edge diffusion of the thermal component through the joint gradient loss. Hence, to have a stable training process, it uses a generator loss function with an adversarial loss ($\mathcal{L}_{adv}$) and the content loss ($\lambda\mathcal{L}_{content}$) defined in \eqref{eq:gan-fm-loss}. 
\begin{equation}\label{eq:gan-fm-loss}
    Loss = \mathcal{L}_{adv} + \lambda \times \mathcal{L}_{content},
\end{equation}
where $\lambda$ is a user-defined hyperparameter.
Unlike typical global discriminators, the Markovian discriminator attempts to discriminate each patch of input images, limiting the network's attention to limited regions and forcing the fused outputs to contain more information. 
The experimental results show that fused output contains a rich set of features even suited for human vision. The authors also report that the model is effective for object detection and image segmentation tasks. 
A similar idea of using GAN for IR and RGB image fusion is investigated in \cite{gao2022dcdr}. Here, Gao~\textit{et~al.} develop a Densely Connected Disentangled Representation Generative Adversarial Network (DCDR-GAN), which separates the content and the modal features of infrared and regular RGB images through disentangled representation, and then fuses them separately. The key difference of \cite{zhang2021gan} form \cite{gao2022dcdr} is the dense connection and the loss function used to train the model. The densely linked structures are used in the content encoders and the fusion decoder to reduce information loss. The objective function in this case contains three constituents, namely reconstruction loss ($\mathcal{L}_{rec}$), modal translation \& cycle reconstruction loss ($\mathcal{L}_{cyc}$), and image fusion loss ($\mathcal{L}_{fus}$) as defined in \eqref{eq:dcdr-gan-loss}. 
\begin{equation}\label{eq:dcdr-gan-loss}
    Loss = \mathcal{L}_{rec} + \mathcal{L}_{cyc} + \alpha \times \mathcal{L}_{fus},
\end{equation}
where $\alpha$ is a user-defined hyperparameter. Although the authors report that their approach achieves excellent visual effects and higher index values compared to the state-of-the-art methods, they do not validate its effectiveness for any vision-related applications, including pedestrian detection.

\begin{figure*}[!ht]
\centering
\includegraphics[trim={0.2cm, 0cm, 0.2cm, 0.1cm}, clip, width=0.8\linewidth]{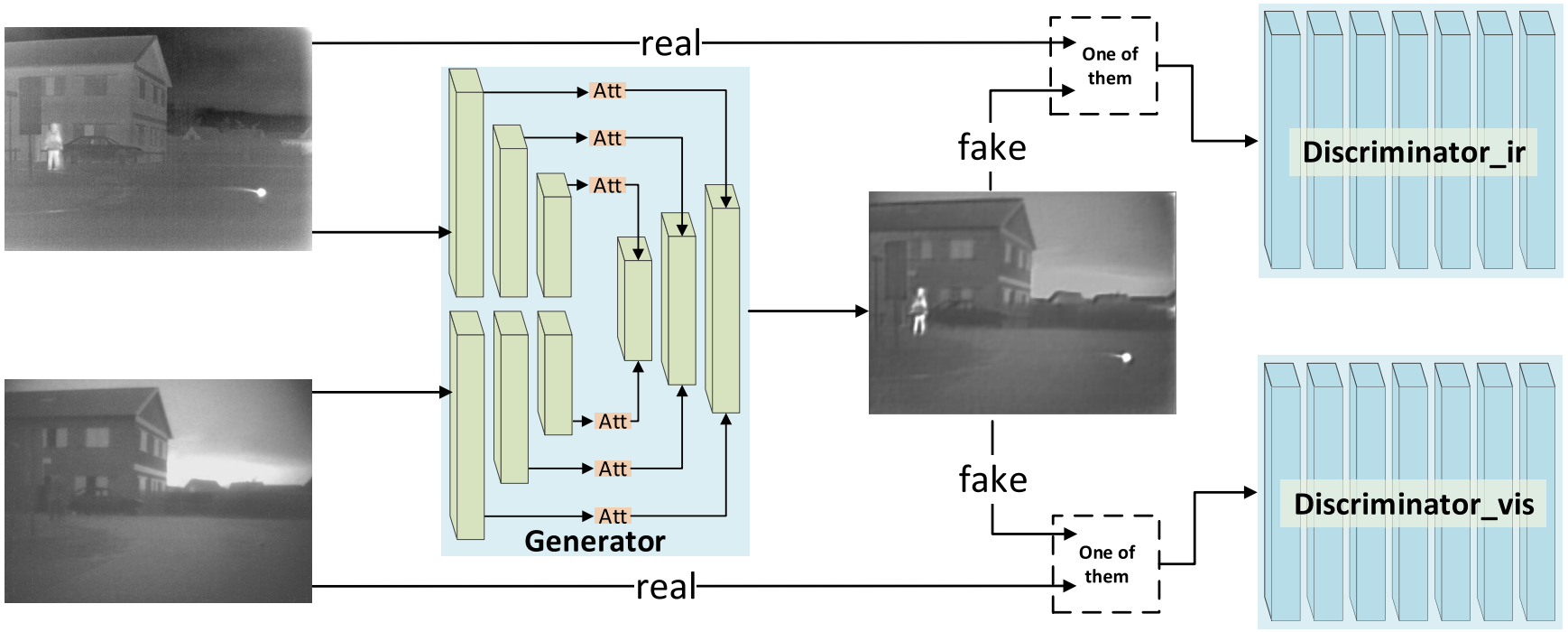}
\caption{Illustration of the MgAN-Fuse model, where ``Att'' refers to the attention module (adapted from \cite{li2020multigrained}). The illumination-aware weighting scheme adaptively estimates the weights for RGB and IR image learning subnetworks. Note: IAN - Illumination aware shallow network that estimates the illumination condition.}
\label{fig:MgANfuse}
\end{figure*}

Yang~\textit{et~al.}~\cite{9335976}, on the other hand, introduce a texture-aware GAN-based fusion technique named conditional generative
adversarial network (TC-GAN). It generates a composite texture map from IR and RGB images for capturing gradient changes in the input scene. This combined texture map is treated as a guide to filter the source inputs, where the filter kernel size is adaptively determined. Finally, they use multiple decision maps and fusion rules to reconstruct texture-enhanced outputs. To preserve the original texture information of the scenes, the gradient loss and adversarial loss are combined to create the generator loss ($Loss_G$) function as in \eqref{eq:TC-GAN-loss}. 
\begin{equation}\label{eq:TC-GAN-loss}
    Loss_G = L_{adv} + \alpha \cdot L_{gra}, 
\end{equation}
where $L_{adv}$ denotes adversarial loss between the generator and the discriminator, and $\alpha$ decides the relative importance of the two functions, which is set to 1,000 in \cite{9335976}. The gradient loss ($L_{gra}$) is computed based on the Mean Squared Error (MSE) loss between the gradient points computed in the visible image, and the combined texture map. The subjective and objective experimental studies show that this texture-aware fusion model can achieve better results than other similar advanced fusion methods. However, the study does not cover the application of the technique for high-level tasks, including pedestrian detection. 
To perceive the discriminative features of the input scenes, researchers of \cite{li2020multigrained} introduce a multi-grained attention mechanism in the generator of a GAN-based fusion (MgAN-Fuse).
It incorporates two encoder modalities with attention heads to encode infrared and regular RGB pictures as depicted in Fig.~\ref{fig:MgANfuse}. The encoder's multi-scale layers produce multi-grained attention maps, which assist the model in focusing on the most discriminative elements so that the fused results not only preserve the foreground targets but also capture context information. 
The decoder, then, concatenates the results of the two encoders to calculate the fused result. Different from other GAN-based fusion approaches, MgAN-Fuse considers a feature loss in the total loss of the generator as defined in \eqref{eq:MgAN-loss}. 
\begin{equation}\label{eq:MgAN-loss}
    Loss_G = L_{adv} + \alpha \cdot L_{con} + \beta L_{fea}, 
\end{equation}
where $Loss_G$, $L_{adv}$, and $L_{fea}$ denote the total loss of the generator, adversarial loss, content loss, and feature loss, respectively. Hence, $\alpha$ and $\beta$ are the control parameters of the three terms set to 1 and 0.1, respectively in \cite{li2020multigrained}.

In \cite{wang2022unidirectional}, the authors provide an incremental improvement to the existing U-shaped encoder-decoder architecture-based thermal and visible image fusion and saliency detection. 
To extract and fuse the detailed semantic information, they perform the fusion of RGB and thermal information at every encoding level, called a local detail-driven fusion module. 
Using these fused features a cross-modal weightage is computed to mine and fuse high-level saliency details that enhance the effectiveness of the features fed to the subsequent encoder level. 
Generally, it is a uni-directional network from encoding concomitant fusion to the final saliency prediction. Their experimental study on various benchmark datasets demonstrates that their model achieves acceptable performance.

\subsection{Pedestrian Detection Techniques}

\begin{figure*}[!t]
\centering
\includegraphics[trim={0cm, 0cm, 10cm, 0cm}, clip, width=0.9\linewidth]{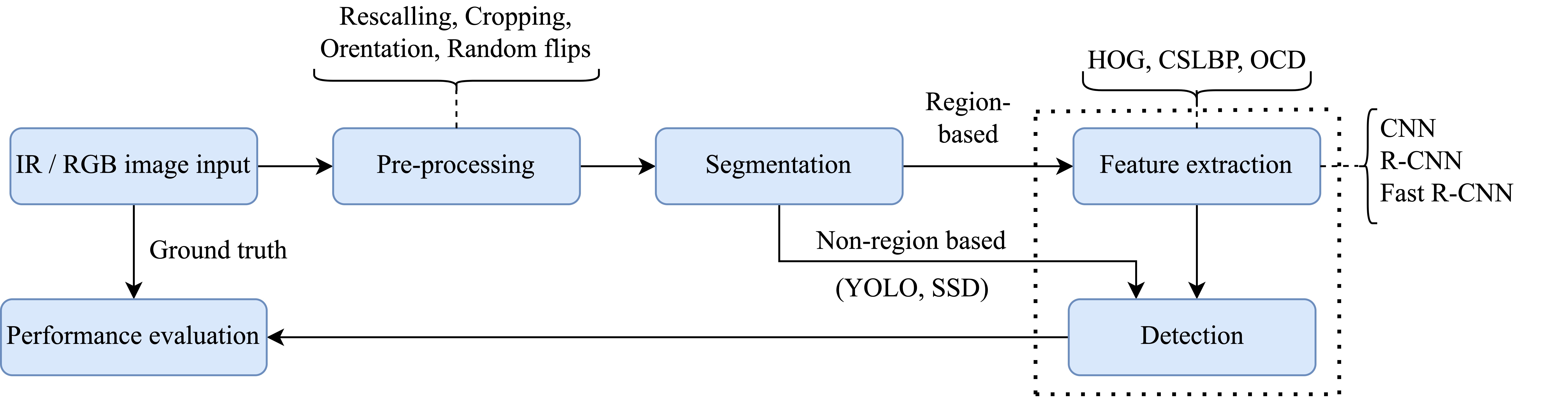}
\caption{Overview of two pedestrian detection approaches-- Region-based (two-stage) technique and non-region-based (single stage) technique.}
\label{pedestrain-flow}
\end{figure*}

\begin{figure*}[!ht]
\centerline{\includegraphics[width=0.955\linewidth]{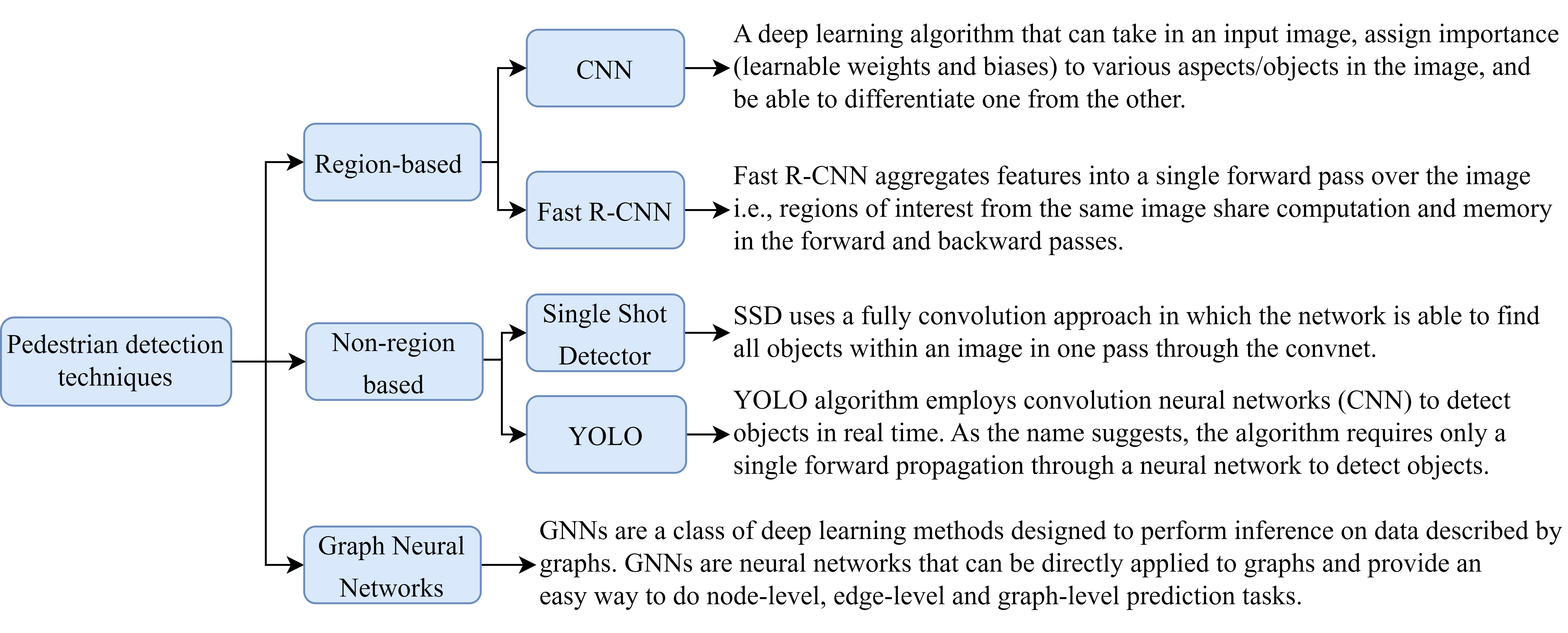}}
\caption{Different machine learning and deep learning techniques implemented for pedestrian detection.}
\label{detectiontechniques}
\end{figure*}

Numerous pedestrian detection techniques have recently been proposed and effectively deployed for practical applications. 
A general pedestrian detection model development pipeline has five key stages as illustrated in Fig.~\ref{pedestrain-flow}: pre-processing, segmentation, feature extraction, classification, and evaluation. 
The pre-processing phase performs operations, such as rescaling the input image to a certain dimension or introducing deformations as part of data augmentation.
The segmentation phase groups pixels based on intensity, color, and other aspects, like texture detail. Wherein, a pixel must abide by a set of predetermined rules to be categorized into regions of pixels that are comparable to one another. 
In the region-based methodologies, the feature extraction phase extracts pedestrian traits 
using different visual features, such as Histogram of Oriented Gradients (HOG), Centre symmetric Local Binary Pattern (CSLBP), and Oriented Chamfer Distance (OCD). Finally, the classification phase aims to identify which candidates within the region correspond to the human form. The classifier provides a binary flag indicating whether a region is positive, i.e., including pedestrians. 
Fig.~\ref{detectiontechniques} presents different pedestrian detection approaches based on their strategies, while table~\ref{tab:detectiontable} summarizes important related works, including their performances, advantages, and drawbacks. 

\onecolumn
{\small
\setlength{\tabcolsep}{2.5pt}
\renewcommand{\arraystretch}{1.8}
\begin{landscape}
\begin{longtable}[c]
{T{1.6cm}T{2.0cm}T{4.5cm}T{4.5cm}T{3.5cm}T{3.5cm}}
\caption{Summary of pedestrian detection approaches}
\label{tab:detectiontable}\\
\hline  
\textbf{Ref.} &
  \centering\textbf{Dataset} &
  \centering\textbf{Methodology} &
  \centering\textbf{Merits} &
  \centering\textbf{Best Performance} &
  \hspace{1cm}\textbf{Limitations}  \\
  \hline
 \endfirsthead

 \hline
 \multicolumn{6}{|c|}{Continuation of Table \ref{tab:detectiontable}}\\
 \hline
 \textbf{Ref.} &
  \centering\textbf{Dataset} &
  \centering\textbf{Methodology} &
  \centering\textbf{Merits} &
  \centering\textbf{Best Performance} &
  \hspace{1cm}\textbf{Limitations}\\
  
  \hline
 \endhead

 \hline
 \endfoot

 \hline
 \multicolumn{6}{| c |}{End of Table \ref{tab:detectiontable}}\\
 \hline\hline
 \endlastfoot
& & & & & \\

Gonzalez \textit{et~al.} \cite{gonzalez2016pedestrian} &
  CVC-14, KAIST Multi-spectral Pedestrian & Fusion of handcrafted features; ML classifiers are built holistically using the whole image as one and patch/part-based models.
  & Introduced a new dataset that includes visible and FIR samples---CVC-14 covering night and day time scenarios &
  RF detector using HOG+LBP on daytime FIR - 16.7\% AMR  &
  The experiments are incomplete. Do not apply the same kind of fusion techniques to both the datasets. \\ \hline
Cadena \textit{et~al.} \cite{cadena2019pedestrian} &
  JAAD dataset &  2D Pose Estimation and Graph Convolutional Networks &
  Greater prediction speed. The advantage of using human key points as it uses a smaller network to recognize actions. &
  Accuracy of 91.94\% in pedestrian crossing prediction. &
  Had a small number of video frames and the algorithm had only been tested on a few samples. \\ \hline
Li \textit{et~al.} \cite{li2019illumination} &  KAIST Multispectral Pedestrian &  Illumination-aware Faster R-CNN (IAF R-CNN). Illumination-Aware Network (IAN) &  Improved the final detection performance in both good and unfavorable lighting circumstances. &  Miss rate - 15.73\%, computation time - 0.21s/image &  - \\ \hline
Konig \textit{et~al.} \cite{konig2017fully} &
  Caltech, CVC-09, KAIST &  Multispectral RPN built up on the pre-trained very deep convolutional network VGG-16 &
  BDT boosted detection performance by lowering the incidence of false positives. &  Fusion RPN alone increased the state-of-the-art by around 18\% with a LaMR of 29.83\% (with the inclusion of the BDT) on the KAIST. &  Resulted in more false positive (FP) detections. \\ \hline
Luo~\textit{et~al.} \cite{luo2020cascade} &  NJUD, STEREO, NLPR, LFSD, SSD, RGBD135, NJU2K &  Cascade Graph Neural Networks (Cas-GNN) &  Used pieces of information from both 2D (color) appearance and 3D geometry (depth) information which enhanced the output for object detection. &  Outperformed all the state-of-the-art models in terms of mean absolute error, the precision-recall curve, F-measure, S-measure, and E-measure. &  - \\ \hline
Chaitanya \textit{et~al.} \cite{devaguptapu2019borrow} &  KAIST and FLIR ADAS &  Fast R-CNN &  Achieved better performance even when trained on 1/4 dataset. &  Surpassed the baseline framework by an increase of 5\% on KAIST and about 8\% on the FLIR dataset. &  Not able to detect the objects when the objects were far from the camera or relatively smaller. \\ \hline
Li~\textit{et~al.} \cite{li2021yolo} &  KAIST and FLIR &  YOLOv5, Path Aggregation Network (PAN), Spatial Pyramid Pooling (SPP) &  Detected objects from infrared images despite the image's low resolution and unclear features. &  Obtained mAP of 83.5\% which was far better than the Faster R-CNN, MMTOD-CG, RefineDet, and TermalDet. &  - \\ \hline
Kang \textit{et~al.} \cite{kang2017t} &  ImageNet VID, YouTubeObjects (YTO) &
  DeepID-Net (an RCNN extension), CRAFT (a Faster R-CNN extension) &
  Improved accuracy. &  Outperformed all the existing methods by 15\% improvement in detection. &  Worked best on only still image object detection. \\ \hline
Wei~\textit{et~al.} \cite{wei2019enhanced} &  KITTI &  Finetuned on the reduced VGG-16 model: pre-trained on the ILSVRC CLS-LOC dataset. &  Reduced false proposals and also detected more small objects. Avoided producing multiple bounding boxes for a single object. &  Network inference speed is only 0.08 seconds, which is much faster than the other top-ranked published methods in KITTI. &  Did not achieve the best accuracy. \\ \hline
Shen \textit{et~al.} \cite{shen2017msr} &  UCID and BSD &  CNN and Retinex theory &  - &  - & As opposed to learning from data, kernel parameters relied on artificial settings, which reduced the model's accuracy and flexibility.\\ \hline
Zhou \textit{et~al.} \cite{zhou2021ast} &  ETH and UCY &  Spatial and Temporal GNNs for modeling interactions. &  Spatial and temporal GNNs employed graph attention to adaptively assign interaction weights among graph nodes. &
  Reduced error by 7\% in ADE measure, and 19\% in FDE measure when compared to the SR-LSTM-2 and Social-STGCNN respectively. &  Performance degraded as the count of AST-GNN layers grew.  \\ \hline
Shi \textit{et~al.} \cite{Shi_2020_CVPR} &  KITTI &  Point-GNN for 3D Object Detection &  Identified several objects in a single shot and had an auto-registration method. &  The average processing time for one sample in the validation split was 643ms. With a score of 93.11, it outperformed the prior models on the Easy level BEV Car detection by 3.45. &  Point sampling and grouping on a wide scale incurred significant computational overhead.  \\ \hline
Wang \textit{et~al.} \cite{wang2019unsupervised} &  ETH and UCY &  GNN for Trajectory Prediction (GNN-TP) model &  Used unsupervised interaction inference based on a GNN for pedestrian trajectory prediction. &  Used ADE and FDE measures to report the performance. &  - \\ \hline
Xu~\textit{et~al.} \cite{Xu_2022_CVPR} &  ETH and UCY &  Transferable GNN (T-GNN) framework &  Used GNN to extract detailed spatial-temporal data representations of features. &  Outperformed the Social-STGCNN and SGCN models on the ADE metric by 21.31\%, and the PCENet and SGCN models on the FDE metric by 20.52\%. & {Accessed only the observed trajectories from the validation set. Reduced performance due to removal of the graph attention component.} \\ \hline
Said \textit{et~al.} \cite{said2019pedestrian} &  Caltech, CVC-09, KAIST  &
  SqueezeNet CNN; Detection - YOLOv2 and Feature extractor - SqueezeNet &
  Real-time detection was possible. Can be deployed to an embedded system or mobile application for an ADAS. &  Inference speed - 32.4 FPS. In terms of mAP, the inclusion of SqueezeNet (75.8\%) was superior to that of YOLOv2+DarkNet (49.1\%) or SSD+SqueezeNet (71.6\%). &
The implemented model was not more accurate (75.8\%) when compared to Fused DNN, which had an accuracy of about 89.05\%. \\ \hline
Zhao \textit{et~al.} \cite{zhao2021graphfpn} &  Microsoft COCO 2017 &
 GraphFPN in the object detection task by integrating it into the Faster R-CNN. &  Local channel attention in a GNN is a perfect fit for big networks like GraphFPN due to its reduced computational cost and greater spatial adaptivity. &  Outperformed state-of-the-art algorithms by at least 1.1\% and had the highest AP (43.7\%). AP is 10.6\% greater than the Faster-RCNN. &
  Construction of superpixel hierarchy with COB costs more time. \\ \hline
Mo~\textit{et~al.} \cite{mo2021graph} &  Next Generation SIMulation (NGSIM) US-101 &  GNN-RNN based model &  Discovered that the prediction accuracy was influenced by both the target vehicle's unique dynamics feature and its interactions with other vehicles. &  Performed better in longer-term prediction than GRIP and CS-LSTM (3-5 sec) in terms of RMSE for STP. &
  Failed to anticipate the lane-change movement that occurred in the next 5 seconds. GRIP performed better in STP for the short-term prediction. \\ \hline
  Du~\textit{et~al.} \cite{du2017fused} &
  Caltech, ETH, TudBrussels &  Fused DNN using SSD &  Used SNF to scale them based on the classification probabilities. &  Achieved around 5\% less false positives than the others. Processing time was 0.3s and 0.16s per image. &  Performance is greatly harmed by hard rejection, and with soft rejection, the detection is poor. \\ \hline
Fang \textit{et~al.} \cite{fang2019tinier} &  PASCAL VOC and COCO &  Tinier YOLO - Originated from Tiny-YOLO-V3 &  Focused on the network performance of the model size, detection speed, and accuracy, to solve issues of Tiny-YOLO-V3. &  Achieved 25 FPS real-time performance and a mAP of 65.7\% on PASCAL VOC and 34.0\% on COCO datasets. &  Performed better at identifying medium and big objects than it does at detecting tiny ones on COCO. \\ \hline
Guo~\textit{et~al.} \cite{guo2021dynamic} &  ExDark &
  Novel framework for the end-to-end training of low-light image enhancement &
  Improved the detection performance of small objects. Reduced the computational cost required to generate region proposals. &  - &
  Described enhancement methods but no specific object detection technique was performed on the detection part. \\ \hline
Brazil \textit{et~al.} \cite{brazil2017illuminating} &
  Caltech and KITTI. &  Faster R-CNN &  Elevated the performance by network diversification by stages. &  Achieved a remarkable MR of 7.36\%. Reached 63.05 mAP score on the moderate setting for the pedestrian class. Run-time - 0.21s/image. &  Much slower than the various models when tested on KITTI. \\ \hline
Lu~\textit{et~al.} \cite{lu2020semantic} &  CityPersons &  Fast RCNN &  Fast and can be used when upscaling the image up to 1.3x. &  Performed best on the R set and outperformed the past+Grid method on HO by using very few iterations. &  Adaptive NMS outperformed HBAN in a sector where it had to discriminate individuals in the crowd by a small margin in R set. \\ \hline
Ghose~\textit{et~al.} \cite{ghose2019pedestrian} &  KAIST Multi-spectral Pedestrian &  Faster RCNN trained on augmented data of raw thermal images with their saliency maps & Detected the false positive items such as trees frequently which boosted accuracy and decreased detection time. &  A miss rate of 30.4\% for daytime and 21\% for nighttime images was achieved, which was an improvement of 13.4\% and 19.4\% over the baseline, respectively. &  Worked better at night time and had a lower miss rate than the daytime. \\ \hline
\end{longtable}

\vspace{-1 pt}
\footnotesize{Note:{  ``-" means no information available.}}
\end{landscape}
}

\subsubsection{Region-based Pedestrian Detection}

\begin{figure*}[!t]
\centering
\includegraphics[trim={0.2cm, 0.2cm, 0.3cm, 0.45cm}, clip, width=0.99\linewidth]{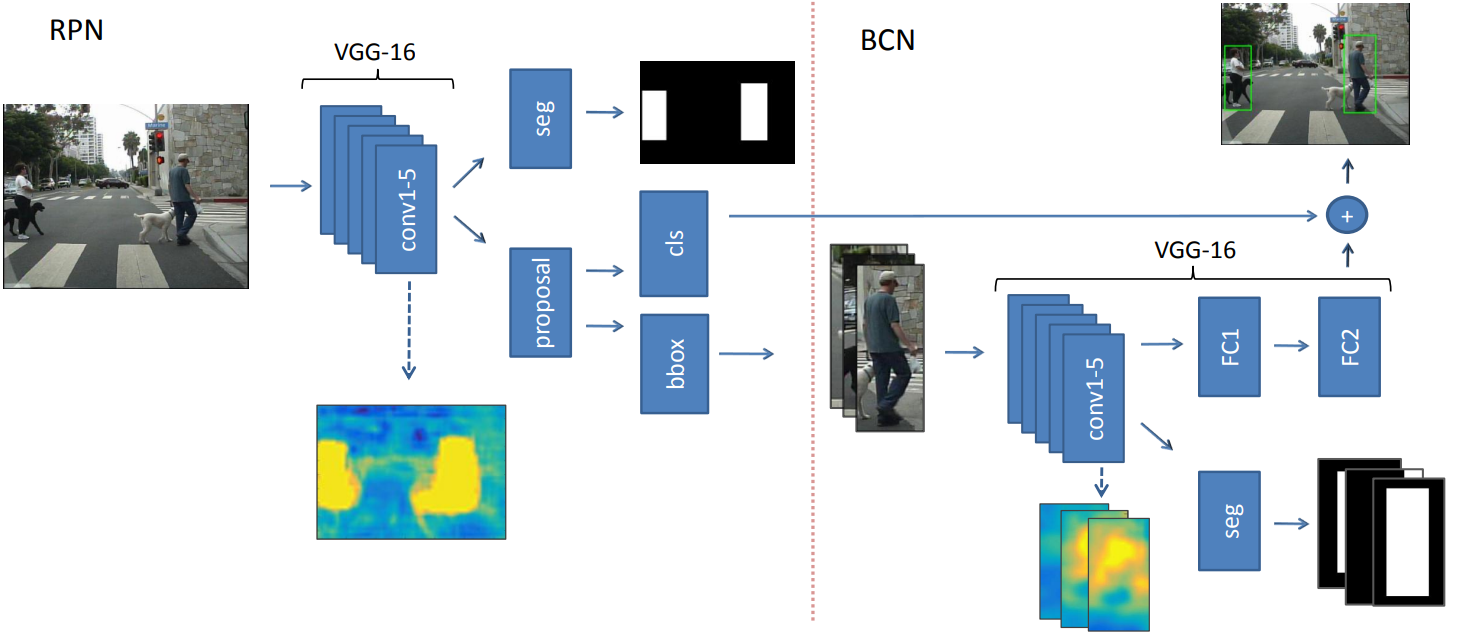}
\caption{Illustration of the SDS-RCNN pedestrian detection model (adapted as it is from \cite{brazil2017illuminating}). It subsumes two modules--(i) a Region Proposal Network (RPN) that generates candidate bounding boxes (BB) and corresponding scores, and (ii) a Binary Classification Network (BCN) that refines the BB scores. The semantic features generated by the segmentation layer are fused into shared conv1-5 layers of each stage.}
\label{fig:SDS-RCNN}
\end{figure*}

Region-based pedestrian detection approaches perform pedestrian detection on automatically selected Regions of Interest (RoI). However, these algorithms face challenges when there is a significant variation in object size and lighting conditions or occluded objects. 
To address these challenges, the authors of \cite{wei2019enhanced} propose deconvolution-based information fusion at low-scale feature maps along with the application of soft Non-Maximal Suppression (NMS) across region proposals at multiple feature scales. The existing region-based detectors use default anchor boxes with predefined sizes to generate object proposals. But in driving environments, like roads and highways, the target object's dimensions can not exceed a certain scale, for instance, the width of a vehicle should not exceed lane width. Thus, to better estimate the anchor box settings, the authors exploit the distributions of the objects' aspect ratio statistics from the KITTI training samples. 
With a negligible inference time overhead, their model shows considerable detection performance improvement compared to the existing algorithms. 
Brazil~\textit{et~al.}~\cite{brazil2017illuminating} improve upon the faster R-CNN~\cite{ren2015faster} by introducing a joint learning framework on semantic segmentation and pedestrian detection (cf. Fig.~\ref{fig:SDS-RCNN}), called Simultaneous Detection and Segmentation R-CNN (SDS-RCNN). Their study shows that the segmentation-infused supervision helps the model to learn information-rich features that are semantically meaningful and robust to shape and occlusion. 

\begin{figure*}[!ht]
\centering
\includegraphics[trim={0.01cm, 0.2cm, 0.01cm, 0.2cm}, clip, width=0.99\linewidth]{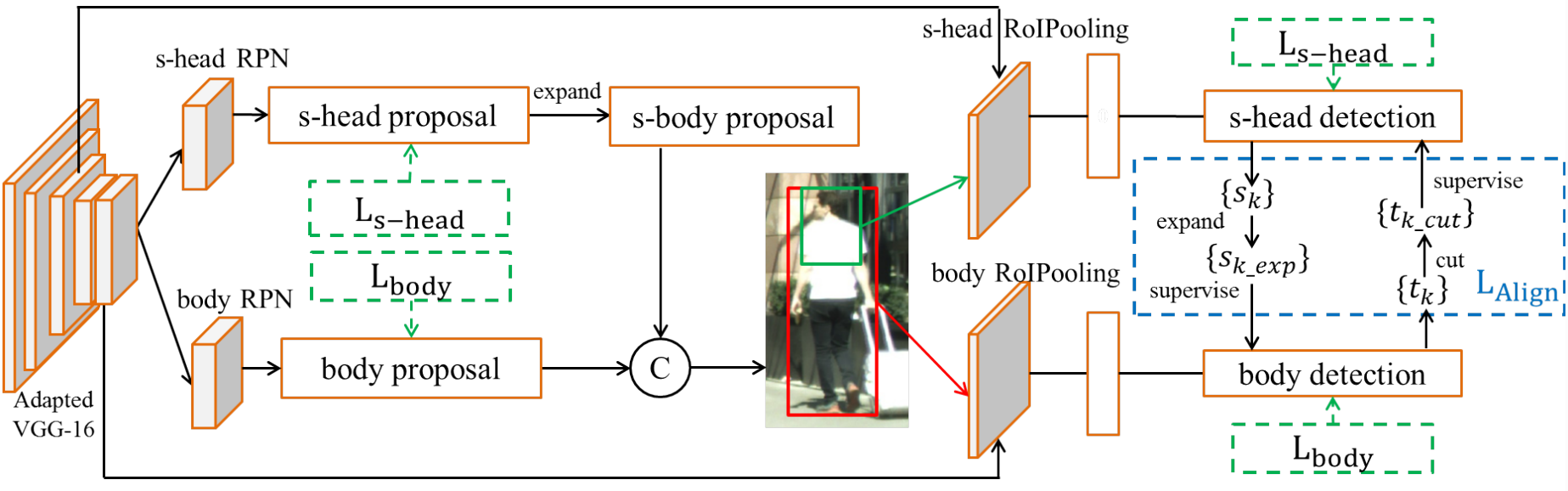}
\caption{Illustration of the HBAN pedestrian detection model (adapted as it is from \cite{lu2020semantic}). It subsumes two RPN subnetworks: (i) for the Semantic Head (s-Head) region proposal and (ii) for the human body region proposal. The s-head and body proposals are aligned and combined. NMS is applied to produce final pedestrian detection results.}
\label{fig:HBAN}
\end{figure*}

Similarly, the work \cite{lu2020semantic} also employs the faster R-CNN as a backbone detector. Its authors hypothesize that the human head region provides a strong cue due to its stable structural appearance, visibility, and relative location to the body. They developed a new architecture---Head-Body Alignment Net (HBAN) (cf.~Fig.~\ref{fig:HBAN}), which has an additional detection module for head detection in parallel with a human body detection module. The Semantic Head (S-head) and body proposals are generated simultaneously by minimizing an alignment loss (AlignLoss) defined in \eqref{eq:AlignLoss}.
\begin{equation}\label{eq:AlignLoss}
    L_{RCNN} = L_{body} + L_{S\text{-}head} + L_{align}, 
\end{equation}
where $ L_{body}$, and $L_{S\text{-}head}$ are similar loss computations used in traditional faster R-CNN representing the classification loss and regression loss of body RoI and s-head RoI wrt their ground truths, respectively. Hence, $L_{align}$ estimated by \eqref{eq:align} makes sure the S-head RoI is located at the top center of the corresponding body RoI of the same pedestrian. 
\begin{equation}\label{eq:align}
    L_{align} = \frac{1}{|B|} \sum_{s\in B}\left( \Delta(t, S_{e}) + \Delta(s, t_{c}) \right), 
\end{equation}
where $t_c$, $t$, $S_{e}$ denote a virtual S-head region cut from the respective target body box or ground truth, the virtual body region expanded from the S-head box ($s$), respectively. Hence, $B$ and $\Delta(\cdot)$
represent a batch of RoIs whose S-head labels are positive and the Smooth-L1 loss, respectively. 
Their experimental study on the CityPersons dataset shows that the HBAN outperforms the state-of-the-art methods in situations, where there is considerable occlusion in the scenes. 



\begin{figure*}[!ht]
\centering
\includegraphics[trim={0cm, 0cm, 0.01cm, 0cm}, clip, width=0.99\linewidth]{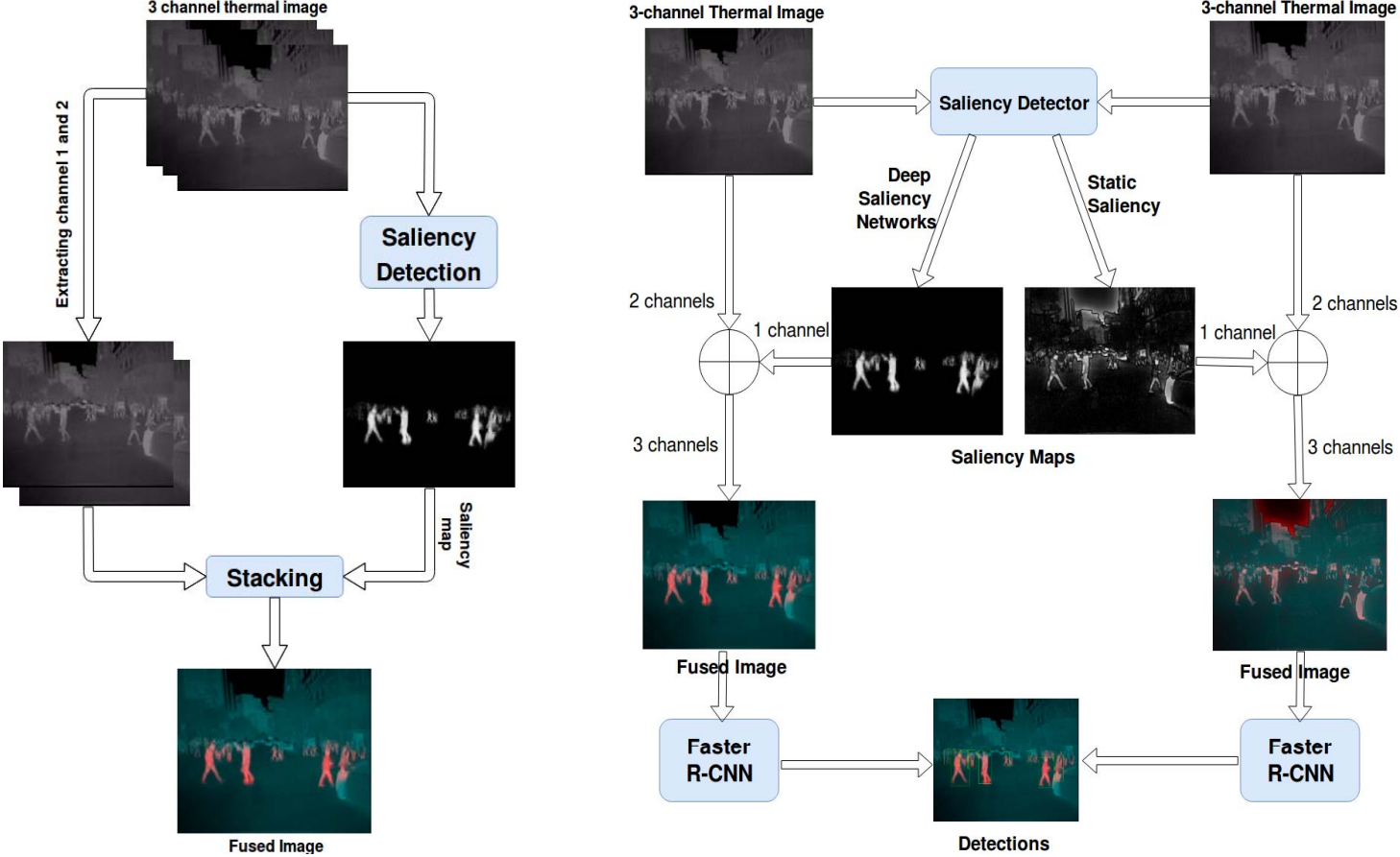}
\caption{Illustration of the thermal image saliency fusion-based pedestrian detection model (adapted as it is from \cite{ghose2019pedestrian}). It subsumes two stages: (i) Augmentation of thermal images with their saliency maps, and (ii) Training a Faster R-CNN on augmented images for pedestrian detection.}
\label{fig:saliency-fuse}
\end{figure*}

\begin{figure*}[!ht]
\centering
\includegraphics[trim={0.1cm, 0cm, 0.1cm, 0cm}, clip, width=0.99\linewidth]{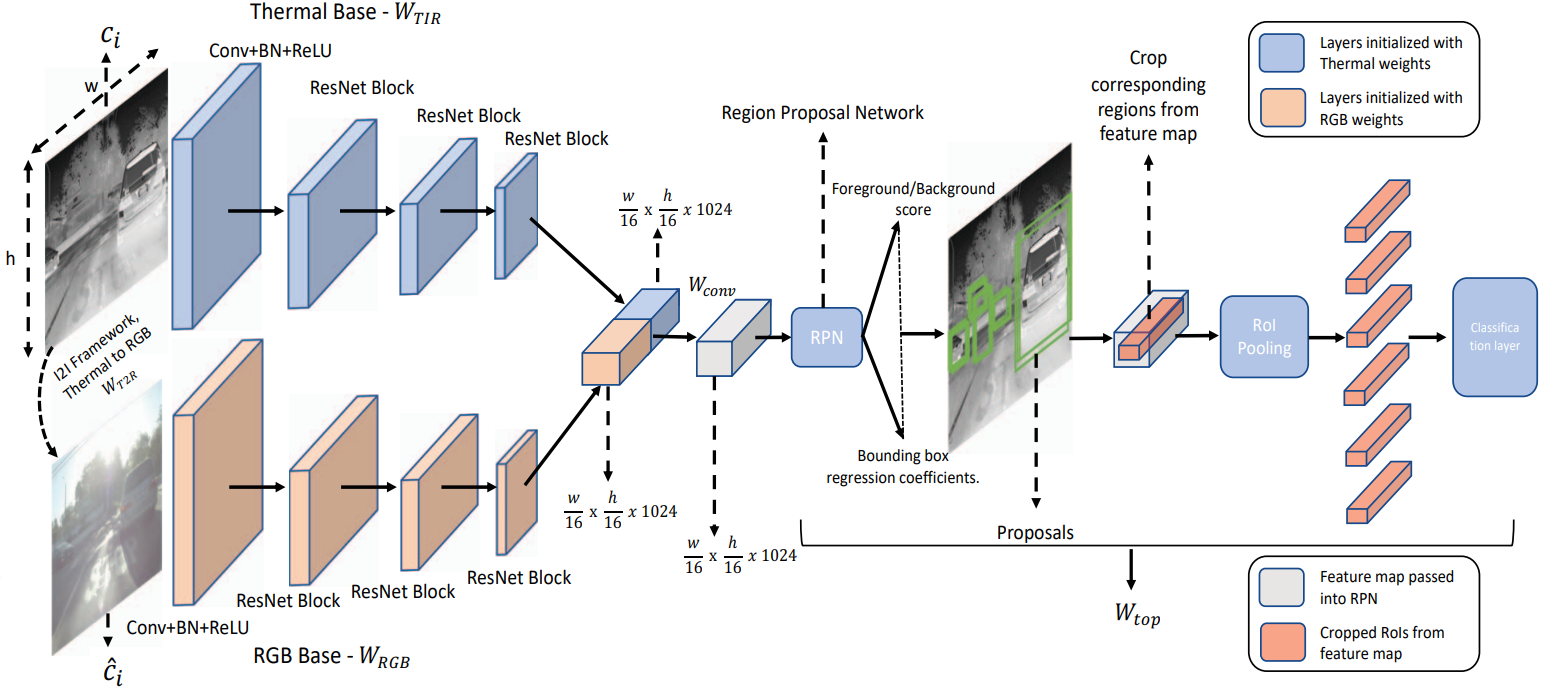}
\caption{Illustration of the pseudo-RGB image-based feature map fusion and object detection model (adapted as it is from \cite{devaguptapu2019borrow}). It subsumes three stages: (i) Pseduo RGB image generation using pre-trained GAN, (iii) Thermal and RGB base feature learning and fusion, and (ii) Faster R-CNN-based object detection. Note: $W_{T2R}$ - Thermal to RGB image-to-imate framework, $W_{RGB}$ - Pre-rained RGB feature learning network, $W_{TIR}$ - Pre-rained thermal infrared feature learning network, $W_{top}$ - Pre-trained thermal top network, and $W_{conv}$ - Randomly initialised $1\time1$ conv weights.}
\label{fig:t2rgb-fuse}
\end{figure*}

Ghose~\textit{et~al.}~\cite{ghose2019pedestrian} have proposed a novel approach based on faster R-CNN. Instead of fusing thermal and visible images, they augment selected raw thermal input channels with their saliency maps. Hence, train an ImageNet pre-trained faster R-CNN from \cite{jjfaster2rcnn}. In this way, they overcome the issue of poor detection performance of thermal images in the daytime. An overview of their approach is shown in Fig.~\ref{fig:saliency-fuse}. The bottleneck of their solution is the algorithm used to generate the saliency map. If the saliency map generator is not robust around the clock, then the results of pedestrian detection become unstable. Their experimental study shows that among various saliency map generators, the proposed model with the $R^3$-Net~\cite{deng2018r3net} achieves the best mAP of 68.5\% during daytime and 73.2\% during nighttime. When compared to a baseline these are 6.9\% and 7.7\% improvements in daytime and nighttime, respectively.
Comparably, Devaguptapu~\textit{et~al.}~\cite{devaguptapu2019borrow} generate pseudo RGB images from raw thermal images using image-to-image (I2I) GAN, like  UNIT \cite{liu2017unsupervised} and CycleGAN \cite{zhu2017unpaired} as illustrated in Fig.~\ref{fig:t2rgb-fuse}. The spatial features from the raw thermal feed and the GAN-generated pseudo visible image are learned through layers of convolution filters and their bottleneck feature maps are depth-wise concatenated and fed to an RPN, eventually performing the object detection operation. 
The fundamental concept of their methodology is to learn from domains with abundant data, like visual (RGB), without explicitly requiring a paired multimodal dataset. 
Experimental analysis of the benchmark datasets namely KAIST~~\cite{hwang2015multispectral} and FLIR~\cite{FLIRDataset-link} reveal that the proposed framework outperforms the existing methods. Overall mAP for a baseline model was 49.39\%, while the authors achieved 53.56\% mAP in \cite{devaguptapu2019borrow} using this multimodal framework. The limitation of this model is overdependent on thermal-to-rgb generator. If the generator produces a noisy RGB counterpart, then the RPN will not be able to propose regions with objections, resulting in a huge toll in processing time or poor object detection performance. 

\subsubsection{Non-region based}
Sasagawa and Nagahara~\cite{sasagawa2020yolo} propose a domain adaption approach that integrates different models trained for distinct visual tasks via generative learning. 
It is a combination of the YOLO and SID, resulting in an improved object detection rate in low-light situations. The experimental analysis on benchmark datasets, like dark COCO~\cite{lin2014microsoft} shows that the integrated model can detect objects in dark images. However, the model is found to be more sensitive than the baseline YOLO model. Thus, it requires further improvements in computational time and object detection accuracy. 
Similarly, The researchers of \cite{li2021yolo} introduce YOLO-FIRI by improving upon another version of the YOLO object detector the YOLOv5~\cite{jocher2020ultralytics} by introducing selected kernel attention mechanism to enhance the feature learning and expanding the cross-stage-partial-connections (CSP) by adding eighteen more layers. Such changes increased the detection of small objects in infrared images. 
According to the authors, when compared to the baseline YOLOv4 a single-stage object detector, the YOLO-FIRI's detection performances are 24.8\%, 26.9\%, 26.0\%, and  21.4\% higher in evaluation metrics indicators AP, AR, and F1, and the mAP50, respectively on KAIST dataset.


\subsubsection{GNN}

\begin{figure*}[!tp]
\centering
\includegraphics[trim={0.1cm, 0cm, 0.1cm, 0cm}, clip, width=0.85\linewidth]{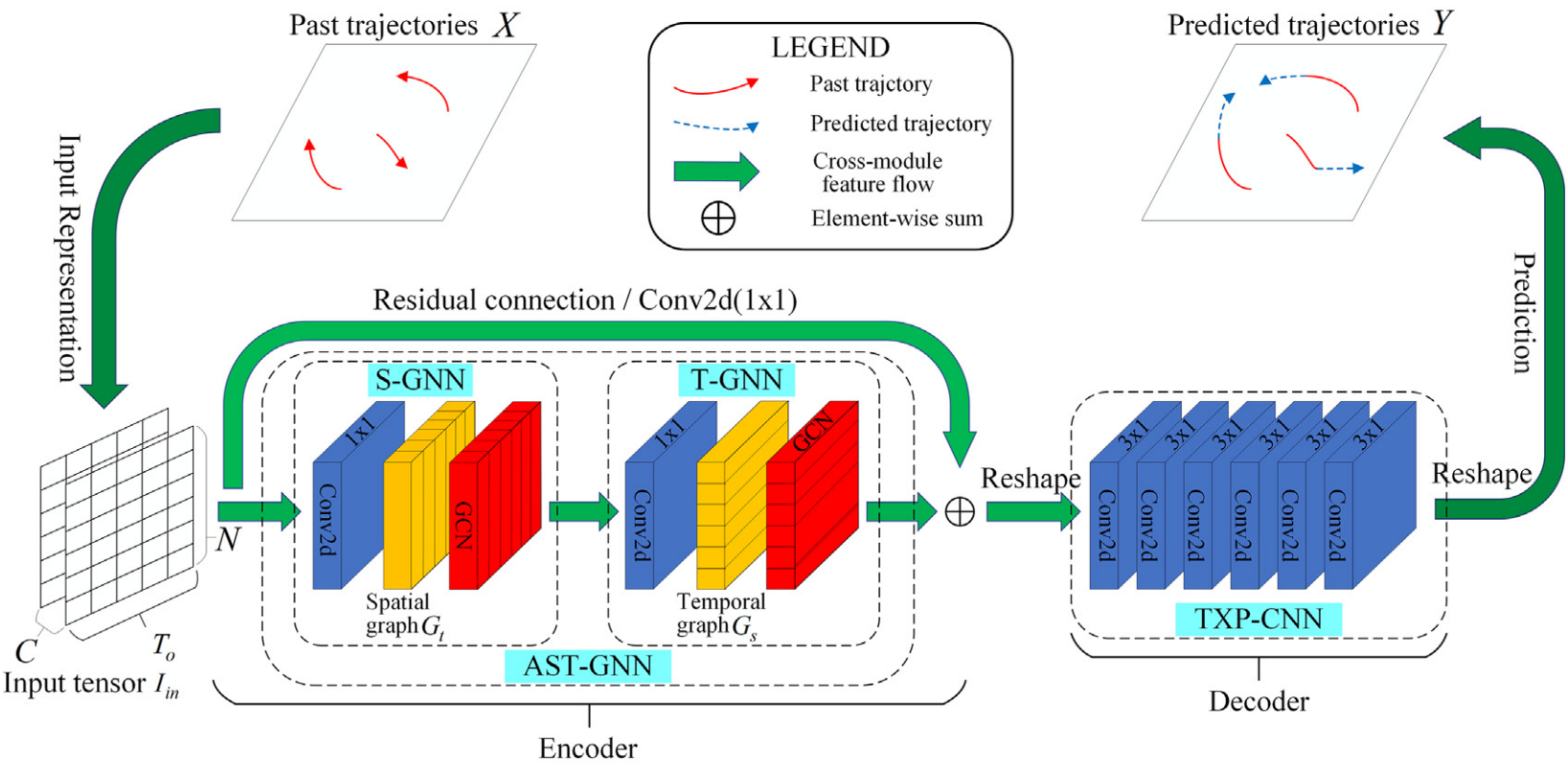}
\caption{The AST-GNN (adapted as it is from \cite{zhou2021ast}). It subsumes three key modules: a spatial graph neural network (S-GNN) for pedestrian interaction learning, a temporal graph neural network (T-GNN) for pedestrian motion information learning, and a time-extrapolator convolutional neural network (TXP-CNN) for predicting future trajectories.}
\label{fig:ast-gnn}
\end{figure*}

Recently GNNs have emerged as powerful representational learning tools that process the input data as graphs~\cite{liu2020towards}. They have been exploited for several high-level computer vision tasks, including pedestrian detection. 
For example, Cadena~\textit{et~al.}~\cite{cadena2019pedestrian} integrate a cascaded pyramid network (CPN) and graph convolutional network (GCN) to predict the intentions of pedestrians crossing (not detection of pedestrian) on urban roads. In which, they represent the pedestrians as non-weighted and undirected graphs of the human body's (joint) key points. 
Similarly, a GNN-based pedestrian trajectory prediction model called GNN-TP is introduced in \cite{wang2019unsupervised}. 
To concurrently infer the relationships between pedestrians and to build an explicit graph structure, it subsumes two modules--- an encoder and a decoder. The encoder module uses observed human trajectories to handle the interaction reasoning problem between pedestrians, while the decoder module learns the dynamical model to predict the pedestrians' trajectories. From an application perspective, the GNN-TP can be used to identify interaction and anticipate direction changes for either a person or a group to avoid collisions. 
Likewise, \cite{zhou2021ast} and \cite{Xu_2022_CVPR} also address the problem of pedestrian trajectory forecasting using GNNs. \cite{zhou2021ast} proposes an integrated attention-based interaction-aware GNN (AST-GNN) that learns spatiotemporal mobility patterns of pedestrians from their past trajectories to predict their future trajectories, inspired by \cite{velickovic2017graph, xu2018powerful, zhao2021stugcn}, as illustrated in Fig.~\ref{fig:ast-gnn}. The model integrates three subnetworks--- a spatial graph neural network (S-GNN) that learns dynamic spatial interactions of the pedestrians, a temporal graph neural network (T-GNN) for building weighted motion features, and a decoder module using an extrapolator convolutional neural network (TXP-CNN) for forecasting trajectories in the temporal domain. When compared to the baseline model \cite{zhao2021stugcn} and other state-of-the-art methods examined in \cite{zhou2021ast}, the  AST-GNN has shown better performance in pedestrian trajectory prediction on the benchmark datasets, such as ETH~\cite{ETHDataset-link}. 

\begin{figure*}[!tp]
\centering
\includegraphics[trim={0.1cm, 0cm, 0.1cm, 0cm}, clip, width=0.95\linewidth]{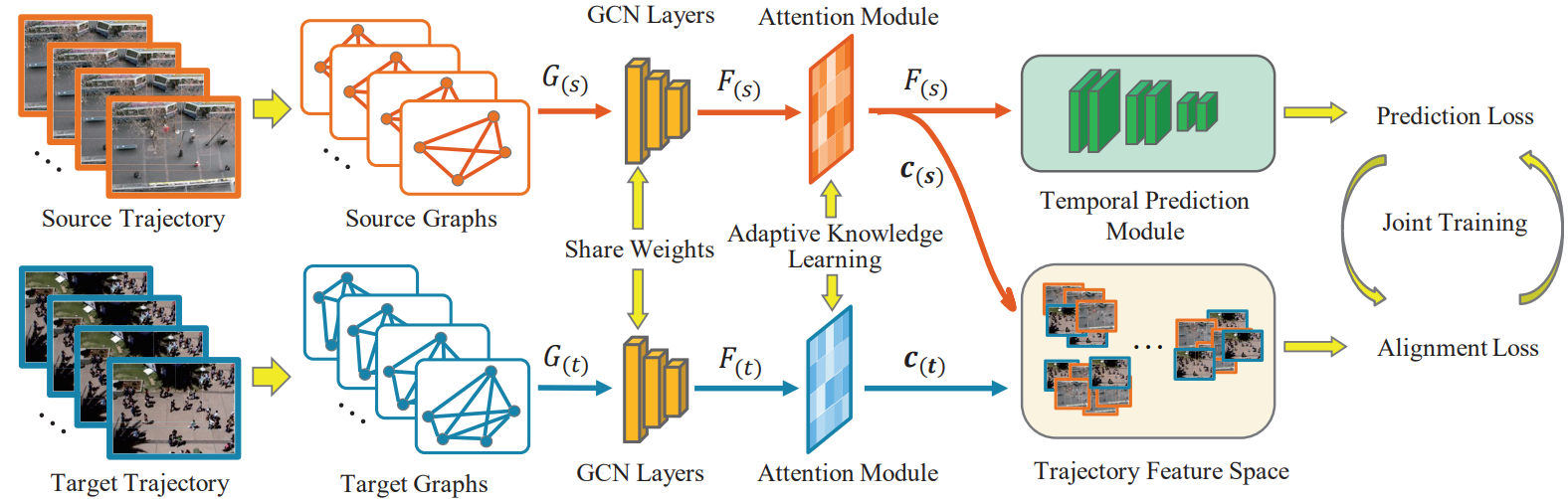}
\caption{Illustration of the T-GNN (adapted as it is from \cite{Xu_2022_CVPR}). It subsumes three sub-modules: i) a GCN that learns spatial-temporal attributes of pedestrians from both source and target domains with shared weights, ii) an attention-based adaptive transfer learning module to extract domain-invariant individual-level features, and iii) a pedestrian trajectory forecaster that is trained by minimizing prediction and alignment losses jointly.}
\label{fig:tgnn} \vspace{-0.5cm}
\end{figure*}

\begin{figure*}[!tp]
\centering
\includegraphics[trim={0.1cm, 0cm, 0.1cm, 0cm}, clip, width=0.99\linewidth]{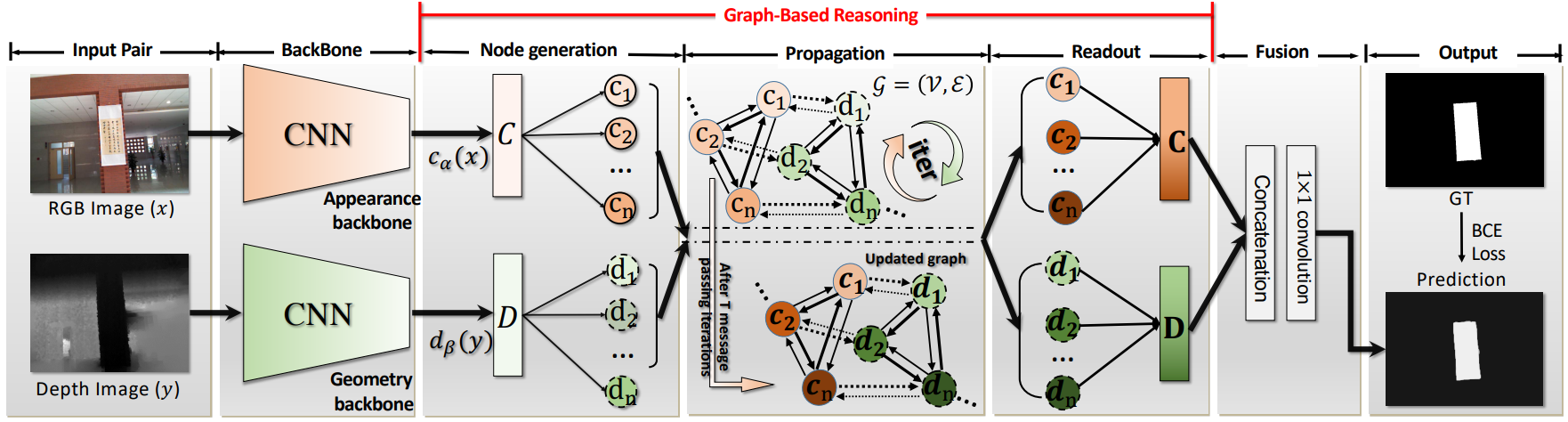}
\caption{Illustration of the Cas-GNN for RGB-D salient object detection (adapted as it is from \cite{luo2020cascade}). It subsumes two separate networks for representation learning from two modalities, namely visible data and depth information. The updated graph features from these modalities are fused to infer accurate salient object regions.}
\label{fig:cas-gnn}
\end{figure*}

\begin{figure*}[!tp]
\centering
\includegraphics[trim={0.1cm, 0cm, 0.1cm, 0cm}, clip, width=0.99\linewidth]{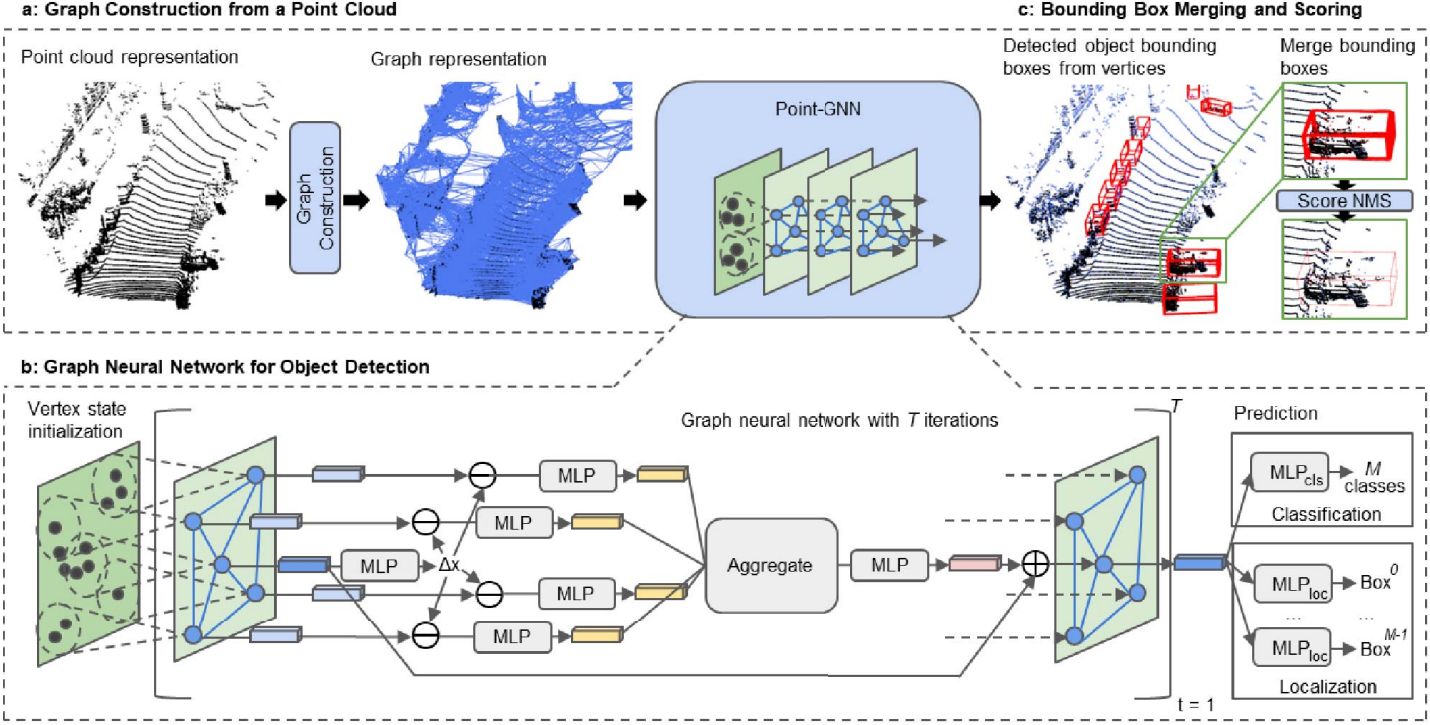}
\caption{Illustration of the point-GNN for object detection in point cloud (adapted as it is from \cite{Shi_2020_CVPR}). It subsumes three modules, viz. (a) graph generator, (b) GNN, and (c) bounding box regressor.}
\label{fig:point-gnn} \vspace{-0.5cm}
\end{figure*}

To address the lack of applicability domain-specific models, Xu~\textit{et~al.}~\cite{Xu_2022_CVPR} develop a transferable graph neural network (T-GNN) that mines domain-invariant features by jointly learning trajectory patterns as well as source and target domain alignments in a unified framework depicted in Fig.~\ref{fig:tgnn}. In this framework, the graph node representations are extracted from respective source and target graphs through a shared weight GCN. Then, an attention-based adaptive mechanism to learn transferable node representations and to align the source and target trajectories. Finally, from the source trajectories, the model learns to forecast the future trajectory by minimizing the prediction and alignment losses, jointly. The authors report that this domain-invariant representation learning model achieves superior performances compared to existing similar approaches on publicly available datasets, like UCY~\cite{lerner2007crowds} and ETH~\cite{ETHDataset-link}.

Besides the above works focusing on pedestrian trajectory prediction using GNN, some researchers attempt to exploit GNN for general object detection tasks. For instance, the authors in \cite{luo2020cascade} approach the salient object problem through a cascade graph neural network (Cas-Gnn), where they learn modality (color and depth)-specific features using separate backbone CNNs and unify the updated graph information from these modalities as shown in Fig.~\ref{fig:cas-gnn} to make the final prediction. To extract multi-scale features from the two modalities it leverages the pyramid pooling module (PPM)~\cite{zhao2017pyramid}, whereby the model overcomes ambiguities in the detection region of objectness. 
Shi and Rajkumar~\cite{Shi_2020_CVPR} on the other hand, build a GNN, named point-GNN that performs object detection on LiDAR point cloud data. Fig.~\ref{fig:point-gnn} depicts their approach consisting of three modules, a graph generator that builds graph representation from the raw point could, a GNN that learns the node feature interaction hierarchically, and a bounding box regressor that identifies the regions with objects, like cars, cyclists, and pedestrians. To handle overfitting issues, the authors use standard data augmentation schemes, such as global rotation, global flipping, box translation, and vertex jitter. It is reported that the point-GNN outperforms existing solutions on the KITTI benchmark~\cite{Geiger2012CVPR} dataset. Although the model achieves a good detection rate for cars and cyclists, it does not seem to perform well for pedestrian detection. Because graph representation of pedestrians is not as accurate as that of the other two objects. In addition, the model's inference speed is also yet to be optimized.   
The afore-discussed works show that there is great potential for using GNNs for object detection and tracking. However, there have been no developments for using them for low-light pedestrian detection tasks at the time of completing this review.

\section{Benchmark Datasets}\label{sec:dataset}

This section provides a succinct synopsis of important benchmark datasets used in the research and development of pedestrian detection models. Table.~\ref{tab:Datasets} summarizes these datasets' details, viz. the number of samples, dimensionality, camera view, lighting condition (day or night), the existing works used the datasets, as well as the best-recorded performances with their evaluation metrics. 

{\small
\setlength{\tabcolsep}{1pt}
\centering
\begin{longtable}[c]
{|T{2cm}|P{1.2cm}|P{1.0 cm}|P{1.95cm}|P{1.0 cm}|T{7cm}|P{1.5cm}|}
\caption{Summary of the benchmark datasets with annotation available for pedestrian detection research, and the existing works that used the datasets}\label{tab:Datasets}\\
\hline \hline
  \centering\textbf{Dataset} &
  \textbf{Count} &
  \textbf{Dim.} &
  \textbf{Camera View} &
  \textbf{Time} &
  \centering\textbf{Notes} &
  \textbf{Existing Works} 
   \\ \hline\hline

OSU \cite{davis2005two}, 2005 &  284 &  $360\times240$ &  Top-view &  Day, Night &  Pedestrian intersection on the Ohio State University campus. &  \cite{davis2005two} \\
\hline

UCY \cite{lerner2007crowds}, 2007 &  18,912 &  640 $\times$ 480 &  Top-view &  Day & Real pedestrian trajectories with rich multi-human interaction scenarios. &  \cite{zhou2021ast} \cite{wang2019unsupervised} \cite{Xu_2022_CVPR} \\
\hline

INO \cite{INODataset-link}, 2007 &  4,200 &  328 $\times$ 254 &  Surveillance &  Day, Night &  Combination of thermal and color images from surveillance video. &  \cite{INODataset-link} \\
\hline

ETH \cite{ETHDataset-link}, 2007 &  2,293 &  640 $\times$ 480 &  Mixed &  Day &  All frames are annotated up to a distance of $\approx 25$ m. $\texthash$ of training frames = 490 with 1,578 annotations, $\texthash$  of test set-1 frames = 999 with 5,193 annotations, $\texthash$  of test set-2 frames = 450 with 2,359 annotations, $\texthash$ of test set-3 frames = 354 with 1,828 annotations. &  \cite{zhou2021ast} \cite{wang2019unsupervised} \cite{Xu_2022_CVPR} \\
\hline

Caltech \cite{CaltechDataset-link}, 2009 &  1,584 &  480 $\times$ 640 &  Driving &  Day & $\approx 250K$ labeled frames, $\approx 132K$ pedestrian frames, $\approx 2300$ unique pedestrians, $\approx 350K$ bounding boxes, and $\approx 126K$ bounding boxes. &  \cite{kim2018pedestrian} \cite{said2019pedestrian} \cite{konig2017fully} \\
\hline

KITTI \cite{Geiger2012CVPR}, 2012 &  14,999 &  384 $\times$ 1280 &  Driving &  Day &  Eight different categories such as cars, pedestrians, and many more. &  \cite{wei2019enhanced} \cite{Shi_2020_CVPR} \cite{chen2022cbi} \cite{brazil2017illuminating} \\
\hline

TNO \cite{TNODataset-link}, 2014 &  5,522 &  768 $\times$ 576 &  Rural/ urban environment &  Night &  Multispectral (intensified visual, near and long-wave infrared or thermal) nighttime images. &  \cite{li2018infrared, zhang2021gan, gao2022dcdr, yang2021infrared, li2020multigrained, li2022mafusion, wang2022res2fusion} \\
\hline
  
KAIST \cite{hwang2015multispectral}, 2015 &  18,410 &  640 $\times$ 480 &  Driving &  Day, Night &  Color and Thermal image pairs in different traffic scenes. & \cite{gonzalez2016pedestrian, golcarenarenji2022illumination, song2021multispectral, kim2018pedestrian, li2019illumination, konig2017fully, deng2021pedestrian, devaguptapu2019borrow, li2021yolo, pei2020fast} \\
\hline

CVC - 14 \cite{CVCDataset-link}, 2016 &  7,785 &  640 $\times$ 512 &  Driving &  Day, Night &  Annotated RGB and FIR day-night pedestrian sequences recorded in the surroundings of Barcelona. &  \cite{gonzalez2016pedestrian} \\
\hline

ExDARK \cite{Exdark}, 2018 &  7,363 &  640 $\times$ 480 &  Mixed &  Day, Night &  Twelve different categories with 609 images for people. &  \cite{xiao2020making} \cite{sasagawa2020yolo} \cite{fisher2021object} \\
\hline

KITTI-dark \cite{rashed2019fusemodnet}, 2019 &  12,919 &  256 $\times$ 1224 &  Driving &  Night &   Synthetically generated samples to represent low-light conditions from the raw samples of KITTI object detection dataset. &  \cite{rashed2019fusemodnet} \\
\hline

FLIR \cite{FLIRDataset-link}, 2020 &  18,449 &  640 $\times$ 512 &  Driving &  Day, Night & It covers challenging environmental scenarios, viz. total darkness, most fog, smoke, inclement weather, and glare.   &   \cite{chen2021exploring} \cite{devaguptapu2019borrow} \cite{li2021yolo} \\
\hline

LLVIP \cite{jia2021llvip}, 2021 &  30,976 &  1080 $\times$ 720 &  Surveillance & Night &  Infrared images and low-light regular-RGB images for pedestrian detection. It uses a thermal infrared wavelength of $8\text{\textendash}14~\mu m$. & \cite{jia2021llvip} \\
\hline\hline
\end{longtable}
}

\subsection{OSU: OSU Thermal Pedestrian Database~\cite{davis2005two}}

This dataset contains a sequence of ten frames totaling 284 images captured in a $240\times360$ pixel resolution using a Raytheon Thermal-Eye 300D infrared camera mounted on a tall building. It includes scenarios of pedestrians walking at intersections and on streets of Ohio State University in varying environmental conditions, such as humidity and weather conditions during day and night. 

\subsection{UCY Dataset~\cite{lerner2007crowds}}

It has more than 18K samples of multi-human interaction in public spaces and their trajectories captured at 2.5 Hz taken from a top view. 

\subsection{INO Video Analytics Dataset~\cite{INODataset-link}}

The outdoor surveillance videos are captured with a permanently mounted VIRxCam (visible and IR camera with long-wave infrared (LWIR) bands) platform throughout the day and in all weather conditions. The videos contain ground-truth annotations of humans/pedestrians.

\subsection{ETH: Caltech Pedestrian Dataset~\cite{ETHDataset-link}} 

This dataset presents videos captured with a forward-looking stereo vision from a moving platform ($\approx 90~cm$ above the ground), driving through a busy pedestrian zone. It also includes challenging conditions, like the multitude of viewpoints, the ambiguity of side vs. semi-frontal views of pedestrians, motion blurs,  low contrast, and weather conditions, including cloudy. In total, there are 2,293 frame sequences containing 10,958 pedestrian annotations. 

\subsection{Caltech: Caltech Pedestrian Dataset~\cite{CaltechDataset-link}} It is one of the most comprehensive standards for pedestrian detection. This dataset is composed of around 10 hours' worth of 640 $\times$ 480 footage that was recorded from a moving car as it traveled through metropolitan traffic, with challenging scenarios, like object occlusion and low resolution. 350,000 bounding boxes as well as 2300 distinct pedestrians were labeled throughout around 250,000 frames (in 137 clips each lasting around a minute).

\subsection{KITTI: Karlsruhe Institute of Technology and Toyota Technological Institute Vision Benchmark~\cite{Geiger2012CVPR}}
It is a demanding benchmark for various real-world computer vision research, including object detection. It has 7481 training and 7518 test images captured from a platform, totaling around 80K labeled objects.
Images typically have a dimension of 384 $\times$ 1280 pixels. This benchmark calculates precision-recall curves for identification using bounding box overlaps and evaluates orientation estimations in bird's eye view using orientation similarity. 

\subsection{TNO Image Fusion Dataset~\cite{TNODataset-link}} 

This dataset is frequently used for the fusion of regular RGB and infrared image-based developments.  
It includes various visual ($390\text{\textendash}700~nm$), near-infrared ($700\text{\textendash}1000~nm$), and long-wave infrared ($8\text{\textendash}12~\mu m$) nighttime images of objects, such as people and vehicles captured in rural and urban environments.

\subsection{KAIST Multispectral Pedestrian Detection Benchmark~\cite{hwang2015multispectral}}

It is a humongous multi-spectral pedestrian dataset widely known as KAIST. It is made up of color and thermal image pairs captured in a range of daylight and nighttime routine traffic situations to account for differences in illumination conditions. It has around 95k color-thermal pairs with each having a dimension of $640\times480$ pixels captured from a moving car. It has more than 100K bounding box annotations for persons, people, and cyclists, wherein 1,182 are unique pedestrians. 

\subsection{CVC-14: Visible-FIR Day-Night Pedestrian Sequence Dataset~\cite{CVCDataset-link}} 
Regular RGB and Far Infrared (FIR) multi-modal video sequences that were captured throughout the day and at night make up this CVC-14 dataset. It provides dedicated training and test sets. The training set contains 3695 daytime images and 3390 nighttime images, while the test set contains 700 samples. In total, there are about 2,000 pedestrians in the daytime and about 1,500 pedestrians in the nighttime.


\subsection{ExDark: Exclusively Dark~\cite{Exdark}}

This dataset has been introduced to support object recognition and image enhancement research. It consists of 7,363 low-light pictures, with twelve object categories having image-level classification labels and local-object bounding box annotations. 
Images are usually in the dimension of $640 \times 480$. 

\subsection{KITTI-Dark~\cite{rashed2019fusemodnet}} 

Using the unsupervised image-to-image translation networks (UNIT)~\cite{liu2017unsupervised}, Rashed~\textit{et~al.}~\cite{rashed2019fusemodnet} generate near-realistic low-lighting autonomous driving (AD) condition images from the KITTI raw dataset, called KTTI-Dark. 
It has 12,919 samples that are the size of $256\times1224$.

\subsection{FLIR: FREE Teledyne FLIR Thermal Dataset~\cite{FLIRDataset-link}}

This dataset focuses on the development of advanced driver-assistive systems (ADAS) using visible and thermal (RGB-T) sensor fusion techniques. 
It consists of 26,442 densely annotated frames with 520,000 bounding box labels for fifteen different objects, like person, bike, car, motorcycle, bus, train, truck, traffic light, fire hydrant, street sign, dog, skateboard, stroller, scooter, and other vehicles. It has mutually exclusive sets of training and validation samples consisting of 9,711 thermal and 9,233 RGB images.

\subsection{LLVIP: Low-light Visible-infrared Paired~\cite{jia2021llvip}}

It includes a total of 30976 images annotated for pedestrians from 15488 visible and IR pairs. The majority of the samples represent very dark lighting conditions.

\section{Conclusion}\label{sec:conclusion}

Detecting pedestrians is always a difficult task, especially in low-light situations. The regular RGB images contain relatively little information that can lead to misinterpretation of pedestrians and other objects. Due to advancements in artificial intelligence and computational resources, pedestrian detection in low-light environments has received industrial-academic interest.  
For example, using the information from thermal, LiDAR, or FIR sensors has been shown to be quite successful in real-world pedestrian detection-based applications. 
This study has reviewed several key articles that focus on the fusion of regular visible spectrum and infrared spectrum-based images. 
The image fusion techniques and pedestrian detection algorithms are systematically categorized and discussed objectively.  
This study also summarizes the important benchmark datasets that have been widely used by the research community in the field of pedestrian detection and tracking. 
The future direction of this study will include benchmarking the pedestrian detection algorithms on resource-limited hardware platforms for practical applications.








\bibliography{sbib}{}
\bibliographystyle{plain}

\end{document}